\documentclass[journal]{IEEEtran}
\ifCLASSINFOpdf
\else
\fi

\usepackage{graphicx}
\usepackage{multirow}
\usepackage{algpseudocode}
\usepackage{algorithmicx,algorithm}
\usepackage{amsmath}
\usepackage{setspace}
\usepackage{lineno,hyperref}
\usepackage{makecell}
\usepackage{color}
\begin{document}
\title{Real-Time High-Performance Semantic Image Segmentation of Urban Street Scenes}

\author{
Genshun~Dong,
		Yan~Yan,~\IEEEmembership{Member,~IEEE},
		Chunhua~Shen,
        and~Hanzi~Wang,~\IEEEmembership{Senior Member,~IEEE
}
\thanks{This work was supported by the National Key R\&D Program
of China under Grant 2017YFB1302400, by the National Natural
Science Foundation of China under Grants 61571379, U1605252,
61872307. (\emph{Corresponding author: Yan Yan.})}
\thanks{G. Dong, Y. Yan and H. Wang are with the Fujian Key Laboratory of Sensing and
Computing for Smart City, School of Informatics, Xiamen University, China (e-mail: gshdong@qq.com).}
 \thanks{C. Shen is with the School of Computer Science, The University of Adelaide, Adelaide, SA 5005, Australia.}}

\markboth{Journal of \LaTeX\ Class Files}%,~Vol.~14, No.~8, August~2015}%
{Shell \MakeLowercase{\textit{et al.}}: Bare Demo of IEEEtran.cls for IEEE Journals}



\maketitle

\begin{abstract}
Deep Convolutional Neural Networks (DCNNs) have recently shown outstanding performance in semantic image segmentation. However, state-of-the-art DCNN-based semantic segmentation methods usually suffer from high computational complexity due to the use of complex network architectures. This greatly limits their applications in the real-world scenarios that require real-time processing. In this paper, we propose a real-time high-performance DCNN-based method for robust semantic segmentation of urban street scenes, which achieves a good trade-off between accuracy and speed. Specifically, a Lightweight Baseline Network with Atrous convolution and Attention (LBN-AA)
 is firstly used as our baseline network to efficiently obtain dense feature maps. Then, the Distinctive Atrous Spatial Pyramid Pooling (DASPP), which exploits the different sizes of pooling operations to encode the rich and distinctive semantic information, is developed to detect objects at multiple scales.
Meanwhile, a Spatial detail-Preserving Network (SPN) with shallow convolutional layers is designed to generate high-resolution feature maps preserving the detailed spatial information. Finally, a simple but practical Feature Fusion Network (FFN) is used to effectively combine both shallow and deep features from the semantic branch (DASPP) and the spatial branch (SPN), respectively. Extensive experimental results show that the proposed method respectively achieves the accuracy of  73.6\% and 68.0\% mean Intersection over Union (mIoU) with the inference speed of 51.0 fps and 39.3 fps on the challenging Cityscapes and CamVid test datasets  (by only using a single NVIDIA TITAN X card). This demonstrates that the proposed method offers excellent performance at the real-time speed for semantic segmentation of urban street scenes.
\end{abstract}


% \fi
\IEEEpeerreviewmaketitle

\section{Introduction}
\IEEEPARstart{S}{emantic} image segmentation is a fundamental but challenging task in computer vision. It aims to provide detailed pixel-level image classification, which amounts to assign semantic labels to each pixel. It is a critical step to achieve deep understanding of different kinds of objects (such as road, human and car) in urban street scenes, and has been widely used in a variety of intelligent transportation systems, such as automotive driving, vehicle safety and video surveillance \cite{kang2011multiband, li2018traffic, chen2019importance, chen2017importance} These systems usually exhibit a strong demand for real-time inference speed and efficient interaction.

%These applications have a strong demand for semantic segmentation algorithms.
%As one of the tasks of scene understanding, semantic segmentation provides the detailed pixel-level image classification. It needs to classify every pixel in the picture,

Early methods for semantic segmentation generally rely on handcrafted features, such as \cite{shotton2008semantic}. However, the performance of these methods is far from being satisfactory. During the past few years, with the significant development of deep learning \cite{krizhevsky2012imagenet, szegedy2015going}, Deep Convolutional Neural Networks (DCNNs) \cite{lecun1998gradient} have been successfully applied to semantic segmentation.
%in various computer vision tasks, such as image classification \cite{krizhevsky2012imagenet, szegedy2015going} and object detection \cite{erhan2014scalable, girshick2015fast}, and pushed the performance of these tasks to astonishing heights.

Since the first prominent work of Fully Convolutional Network (FCN) \cite{long2015fully}, semantic segmentation has made remarkable progress with DCNNs \cite{chen2014semantic, zhao2017pyramid, chen2018deeplab}. In particular, benefiting from the emergence of a variety of excellent DCNN frameworks, several DCNN-based methods have shown promising results on public benchmark datasets.
%For example,
%Chen et al. [???] propose the DeepLab method, which takes advantage of atrous convolution \cite{holschneider1990real, papandreou2015modeling, giusti2013fast} and atrous spatial pyramid pooling (ASPP) for semantic image segmentation.
%Atrous convolution is beneficial to significantly enlarge the area of receptive fields and obtain dense feature maps. ASPP can be used to robustly segment the objects at multiple scales.
%DeepLab achieves 70.4\% mean IoU (intersection over
%union) on the popular benchmark Cityscapes dataset.
For example, the current state-of-the-art DeepLabv3+ \cite{chen2018encoder} (based on the Xception \cite{chollet2017xception} model) and PSPNet \cite{zhao2017pyramid} (based on ResNet \cite{he2016deep}) respectively achieve about 82\% and 81\% mean Intersection over Union (mIoU) on the Cityscapes dataset \cite{cordts2016cityscapes}.

%. DCNNs have also been introduced into semantic image segmentation
 %However, the results are not satisfactory.
 %After the first prominent work Fully Convolutional Network (FCN) \cite{long2015fully} where the network was trained in an end-to-end and the fully connected layers were replaced by the convolution layers was proposed, DCNNs become the preferred approach for semantic segmentation. Semantic segmentation has made remarkable progress in recent years with DCNNs \cite{long2015fully, chen2014semantic, zhao2017pyramid, chen2018deeplab, yang2018denseaspp, zhang2018context, wu2016wider}. Benefiting from the emergence of a variety of excellent convolutional neural network frameworks, the performance of semantic segmentation achieves very good results on various famous benchmark datasets. For example, the current state-of-the-art DeepLab V3+ \cite{chen2018encoder} and PSPNet \cite{zhao2017pyramid} based on Xception \cite{chollet2017xception} and ResNet \cite{he2016deep} achieve about 82\% and 81\% mean IoUs \cite{cordts2016cityscapes} respectively on the popular benchmark dataset Cityscapes \cite{cordts2016cityscapes}.

\begin{figure}[!t]
\centerline{
\includegraphics[scale=0.33]{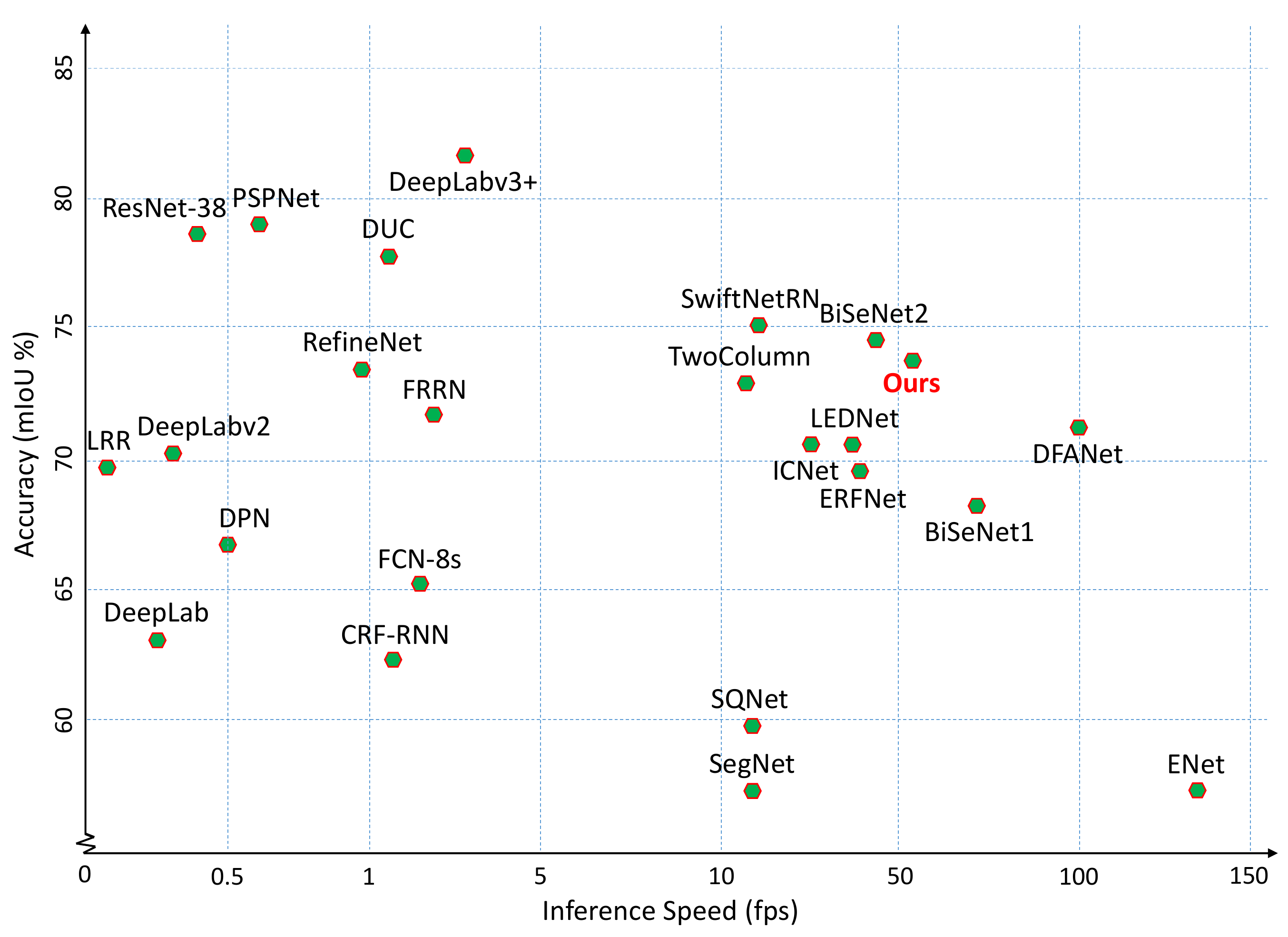}}
\caption{The accuracy (mIoU) and inference speed (fps) obtained by several state-of-the-art methods on the Cityscapes test dataset.} %which are trained using only fine data.
\label{fig:more_more}
\end{figure}

It is common belief that the increase of accuracy almost implies more computational operations and higher memory consumption, specifically for pixel-level classification tasks, such as semantic segmentation. To illustrate this dilemma, Fig.~\ref{fig:more_more} gives the accuracy (mIoU) and inference speed (frames per second (fps)) obtained by several state-of-the-art methods, including FCN-8s \cite{long2015fully}, CRF-RNN \cite{zheng2015conditional}, DeepLab \cite{chen2014semantic}, DeepLabv2 \cite{chen2018deeplab}, DeepLabv3+ \cite{chen2018encoder}, ResNet-38 \cite{wu2016wider}, PSPNet \cite{zhao2017pyramid}, DUC \cite{wang2018understanding}, RefineNet \cite{lin2017refinenet}, LRR \cite{ghiasi2016laplacian}, DPN \cite{liu2015semantic}, FRRN \cite{pohlen2017fullresolution}, TwoColumn \cite{wu2017real}, SegNet \cite{badrinarayanan2015segnet}, SQNet \cite{treml2016speeding}, ENet \cite{paszke2016enet}, ERFNet \cite{romera2018erfnet}, ICNet \cite{zhao2017icnet}, SwiftNetRN \cite{orsic2019defense}, LEDNet \cite{wang2019lednet}, BiSeNet1 \cite{yu2018bisenet}, BiSeNet2 \cite{yu2018bisenet}, DFANet \cite{li2019dfanet} and our proposed method, on the Cityscapes test dataset. Clearly, how to achieve a good tradeoff between accuracy and speed is still a challenging problem.

In general, most of the current state-of-the-art semantic segmentation methods are based on complicated baseline networks (such as VGG \cite{simonyan2014very} or ResNet \cite{he2016deep}), which are originally designed for multi-class classification. These methods usually have thousands of channels and up to hundreds of layers, which are very large in both width and depth. Therefore, although these methods can achieve superior accuracy, the computational complexity is significantly high due to the use of sophisticated DCNN models. For example, the high-accuracy semantic segmentation methods (e.g.,
%DeepLabv3+ \cite{chen2018encoder} and
PSPNet \cite{zhao2017pyramid}) take more than one second to predict a high-resolution image ($1024\times2048$) on a high-end GPU card (e.g., NVIDIA TITAN X). The high computational cost of these methods seriously limits their applications for understanding urban street scenes in intelligent transportation systems (which have a strong demand for inference speed to achieve fast interaction and response). %\cite{he2015convolutional}.

%which is one of the most expensive direction in computer vision. To illustrate, we show this in Figure \ref{fig:more_more}. Most of the current semantic segmentation approaches are based on the wider and deeper high-performance baseline networks \cite{wu2016wider}, such as VGG \cite{simonyan2014very}, ResNet \cite{he2016deep} or Inception \cite{szegedy2017inception} etc, which are designed for multi-class classification. All of these models are quite large and complex, which usually have thousands of channels and hundreds of layers. Although these semantic segmentation networks can achieve higher accuracy, their effectiveness heavily depends on the sophisticated models with huge numbers of parameters and long inference time. This results in a very long processing time for an input image, and many networks need a second or more to process the input.  However, many real-world applications are sensitive for time and have a high demand for inference speed to achieve fast interaction and response. It is especially problematic

% Work in this area is equally important since it can make many practical applications, such as automatic driving or mobile computing etc, feasible, where running time is a critical factor.
%
Compared with the rapid development of high-accuracy semantic segmentation,
it is still left far behind to perform fast (e.g., real-time) semantic segmentation without sacrificing too much segmentation accuracy.
Although several methods \cite{howard2017mobilenets, liu2016ssd,redmon2018yolov3} have been developed to improve the computational efficiency of DCNNs, these methods mainly focus on image classification or object detection. Recently, due to the increasing demand for real-time inference, fast semantic segmentation begins to attract much attention.
Among these fast semantic segmentation methods, SegNet \cite{badrinarayanan2015segnet}, SQNet \cite{treml2016speeding} and ENet \cite{paszke2016enet} are the representative ones.
These methods are able to deal with the full resolution images at the real-time speed. However, the segmentation accuracy of these methods is significantly lower than that of the state-of-the-art high-accuracy methods (the mIoUs obtained by these methods are lower than 60\%). Therefore, it is critical to maintain a reasonable balance between accuracy and speed.
%
%due to the increasing demand for fast inference, many real-time semantic segmentation approaches based on convolutional neural networks have been proposed in recent years. Among these fast semantic segmentation approaches, SegNet \cite{badrinarayanan2015segnet}, ENet \cite{paszke2016enet} and SQNet \cite{treml2016speeding} are representative. These fast semantic segmentation approaches process the full resolution images of the Cityscapes  dataset at real-time speed, but with significantly reduced accuracy compared with the state-of-the-art one, where the final mean IoUs are even lower than 60\%. The above approaches

%In this paper, we propose a practically fast semantic segmentation network framework with excellent prediction results.

In this paper, we propose an efficient and effective method designed for fast inference and high accuracy for semantic segmentation. Unlike previous methods that only consider accuracy or speed, we make a comprehensive tradeoff on these two seemingly contradictory aspects. In particular, instead of relying on the complex DCNN models, we adopt a Lightweight Baseline Network (i.e., the modified MobileNetV2 \cite{sandler2018inverted}) with Atrous convolution \cite{holschneider1990real} and Attention \cite{chen2017sca} (LBN-AA),
%} based on the modified MobileNetV2 \cite{sandler2018inverted},} %()
which requires little memory and a small amount of parameters to achieve fast inference and comparable accuracy. To effectively deal with the multi-scale problem of semantic segmentation, we develop the Distinctive Atrous Spatial Pyramid Pooling (DASPP) to capture objects and context at multiple scales. Meanwhile, a Spatial detail-Preserving Network (SPN) with shallow convolutional layers is designed to preserve affluent spatial details. Finally, a simple but practical Feature Fusion Network (FFN) is used to effectively combine both shallow and deep features so as to obtain the final segmentation results.

Our main contributions are summarized as follows:
\begin{itemize}
\item LBN-AA is adopted as our baseline network, which is able to not only provide sufficient receptive fields, but also obtain dense feature maps in a low computational cost. Moreover, DASPP is proposed to robustly segment objects at multiple scales. Compared with ASPP \cite{chen2017rethinking}, the features obtained by DASPP are richer and more distinctive by taking advantage of the different sizes of pooling operations and neighboring information of each pixel.
\item SPN and FFN are respectively adopted to further improve the accuracy of our method without much loss of speed. SPN is able to accurately preserve the rich spatial information to remedy the loss of spatial details in deep layers, while FFN effectively combines both shallow and deep features from the semantic branch (DASPP) and the spatial branch (SPN), respectively.
\item The above components (i.e., LBN-AA, DASPP, SPN and FFN) are tightly combined and jointly optimized in an integrated network to effectively improve the accuracy and speed of semantic segmentation. The proposed method respectively obtains the accuracy of 73.6\% and 68.0\% mIoU at the inference speed of 51.0 fps and 39.3 fps on the challenging Cityscapes and
CamVid \cite{brostow2008segmentation} test datasets by only using a single NVIDIA TITAN X card. Therefore, the proposed method provides a good balance between accuracy and speed for semantic segmentation of urban street scenes.
\end{itemize}

%\textcolor{red}{In summary, we propose a novel real-time high-performance semantic image segmentation method by effectively combining four components (i.e., LBN-AA, DASPP, SPN and FFN).Experimental results on the challenging Cityscapes and CamVid test datasets have validated the excellent performance of our method in both segmentation accuracy and running speed.}
%Previous works about improving computational efficiency of networks mainly focus on image classification and object detection. Little studies have been targeted towards the fast semantic segmentation. Building a computationally efficient semantic segmentation network is still an open research direction. Contrary to the rapid development of high-accuracy semantic segmentation, it is still backward to make semantic segmentation run faster without sacrificing too much segmentation quality. Therefore, real-time semantic segmentation began to attract everyone attention recently \cite{badrinarayanan2015segnet, paszke2016enet, zhao2017icnet}. Work in this area is equally important since it can make many practical applications, such as automatic driving or mobile computing etc, feasible, where running time is a critical factor.

The remainder of this paper is organized as follows. In Section \ref{relatedwork}, we review the related work. In Section \ref{preliminary}, we give some preliminary knowledge of the proposed method. In Section \ref{proposedmethod}, we explain the proposed method in detail. In Section \ref{experiment}, we present and discuss the experimental results. In Section \ref{conclusion}, we draw the conclusions.%Section \ref{sec:related_work} reviews the related work of FER. Section \ref{sec:proposed_method} firstly gives an overview of our proposed method and then describes the details of the joint loss and the loss weights for loss balancing. The experimental results are given in Section \ref{sec:experiment} and the conclusions are made in Section \ref{sec:conclusion}.

\section{Related Work}
\label{relatedwork}
Traditional semantic segmentation methods (such as Boosting \cite{shotton2009textonboost} or Random Forests \cite{shotton2008semantic}) mainly rely on hand-crafted features to learn the robust representation. However, the results obtained by these methods are still far from being satisfactory. In recent years, the DCNN features
based semantic segmentation methods have made great progress.
%Various methods achieve the state-of-the-art performance in different benchmark datasets, which push semantic segmentation to a higher level.
In the following, we will introduce the high-performance semantic segmentation methods and real-time semantic segmentation methods, respectively. In addition, we also introduce several lightweight DCNN methods related to our proposed method.

\subsection{High-performance Semantic Segmentation Methods}
%One of the most important DCNN-based  in semantic segmentation is the Fully Convolutional Network (FCN). FCN

Long et al. \cite{long2015fully} propose the Fully Convolutional Network (FCN) method, which is a pioneer work to apply DCNN to the semantic segmentation task.
FCN replaces the last fully connected layers of DCNN with the convolution layers so as to take inputs of arbitrary size.
%To refine the segmentation results, FCN fuses the outputs with the feature maps from shallower layers using skip connections \cite{hariharan2015hypercolumns, badrinarayanan2015segnet}.
Currently, most high-performance semantic segmentation methods follow the principle of FCN.

DeepLab \cite{chen2014semantic} introduces the atrous convolution into semantic segmentation, which can effectively expand the receptive fields of the networks.
%The idea of atrous convolution is originally designed for the efficient computation of the undecimated wavelet transform .
Compared with the standard convolution, atrous convolution \cite{holschneider1990real} can obtain denser feature maps without increasing the overall number of parameters and the amount of calculation. To segment objects at multiple scales, DeepLabv2 \cite{chen2018deeplab} further proposes the Atrous Spatial Pyramid Pooling (ASPP). ASPP utilizes multiple parallel atrous convolution layers with filters at different sampling rates to capture objects as well as context at multiple scales.

The encoder-decoder structure is widely used in semantic segmentation methods, such as U-Net \cite{ronneberger2015u} and DeepLabv3+ \cite{chen2018encoder}.
In this structure, the decoder is able to restore the high-resolution feature maps from the low layers by taking advantage of skip connections.
Similar to image pyramid \cite{adelson1984pyramid}, some methods (such as \cite{chen2016attention, zhao2017pyramid}) also use multi-scale feature aggregation to capture different levels of context information \cite{mottaghi2014role}.
Zhao et al. \cite{zhao2017pyramid} append multi-scale pooling layers to improve the information flow between local and global contexts.
In \cite{lin2017refinenet}, a novel multi-path refinement network is proposed to combine multi-scale feature maps in a cascade manner. To refine the outputs, most methods (e.g., \cite{chen2014semantic, zheng2015conditional}) also utilize probabilistic graphical models, such as Conditional Random Fields (CRF) or Markov Random Fields (MRF) {\cite{krahenbuhl2011efficient, zhou2018fc}}, as the post-processing steps to improve the localization of object boundaries.

Most of the above methods can achieve high accuracy on a few benchmark datasets. However, these methods usually suffer from high computational complexity (requiring either a large number of parameters or large-scale floating point calculations, or both).

\subsection{Real-time Semantic Segmentation Methods}
Real-time semantic segmentation methods aim to generate high-quality segmentation results in real-time. In fact, speed is one important factor for
analyzing urban street scenes in intelligent transportation
systems.
%The real-time object detection methods, such as SSD \cite{liu2016ssd} and YOLO \cite{redmon2018yolov3}, have made considerable progress. For example, YOLO is able to perform object detection at a speed of 30 fps with state-of-the-art accuracy  on the COCO test dataset \cite{lin2014microsoft} using a single NVIDIA TITAN X card.

In recent years, real-time semantic segmentation has drawn increasing attention.
Paszke et al. \cite{paszke2016enet} propose the Efficient Neural Network (ENet), which uses a compact encoder-decoder architecture to achieve real-time semantic segmentation. Due to the efficiency of ENet, it can be used for the tasks requiring low latency operations. Efficient Spatial Pyramid Network  (ESPNet) \cite{mehta2018espnet} and Efficient Residual Factorized Network (ERFNet) \cite{romera2018erfnet} are another two efficient real-time semantic segmentation methods, which are faster and more accurate than ENet using the similar number of parameters. In particular, ESPNet makes use of the Efficient Spatial Pyramid module (ESP), which follows the convolution factorization principle that decomposes a standard convolution into a point-wise convolution
and a spatial pyramid of atrous convolutions.
ERFNet proposes an efficient convolutional block, which takes advantage of residual connections and factorized convolutions to balance the trade-off between accuracy and efficiency.
Segmentation Network (SegNet) \cite{badrinarayanan2015segnet} adopts a small encoder-decoder network architecture and memorizes the max-pooling indices to achieve fast inference speed. Although these methods are able to generate the segmentation results in real-time, their accuracy is still not satisfactory.

Recently, Image Cascade Network (ICNet) \cite{zhao2017icnet} utilizes a simplified version of Pyramid Scene Parsing Network (PSPNet) \cite{zhao2017pyramid} and a cascade framework with three branches to efficiently process the high-resolution images.
%ICNet can achieve the speed of 30.3 fps on the $1024\times2048$ images, and 67\% mean IoU on the CityScapes validation dataset. For the test dataset, ICNet can even obtain 69.5\% mean IoU using additional training data from the validation dataset.
ICNet obtains fast inference speed and good accuracy for high-resolution inputs. However, it is not competivie for lower-resolution input images \cite{nekrasov2018light}. Lightweight Encoder-Decoder Network (LEDNet) \cite{wang2019lednet} employs an asymmetric encoder-decoder architecture for real-time semantic segmentation. The encoder adopts two new operations (i.e., channel split and shuffle) in each residual block to significantly reduce the computational cost, while the decoder uses an attention pyramid network to further reduce the network complexity.
Bilateral Segmentation Network (BiSeNet) \cite{yu2018bisenet} designs the spatial path and the semantic path to respectively preserve the spatial information and obtain the sufficient receptive field for improving the accuracy and speed of semantic segmentation.
 Based on the two paths, a new Feature Fusion Module (FFM) is developed to combine the features efficiently.
Although these methods make a better tradeoff between accuracy and speed, there is still a big room for further improvement.

%\textcolor{red}{In this paper, we also propose a real-time semantic segmentation method. However, different from ICNet that incorporates multi-resolution branches, the proposed method achieves an excellent balance of accuracy and speed by effectively exploiting the information from both the semantic branch and the spatial branch.}

\subsection{Lightweight Classification Networks}
In real-world applications, it is usually required that the semantic segmentation methods can obtain the best accuracy with only a low computational cost under some specific platforms (e.g., mobile platform) and application scenarios (e.g., autonomous driving).
This motivates the research towards an optimal balance between accuracy and efficiency. For the past several years, tuning the deep neural architectures to lightweight networks has been an active research area. Representative lightweight networks include ShuffleNet \cite{zhang2018shufflenet}, MobileNetV1 \cite{howard2017mobilenets} and MobileNetV2 \cite{sandler2018inverted}.

ShuffleNet utilizes the novel pointwise group convolutions and channel shuffle operations to effectively reduce the computational cost while maintaining the good accuracy. The shuffle operation facilitates the flow of cross-group information between multiple group convolution layers.
MobileNetV1 is specifically designed for mobile devices and embedded applications, and it is mainly composed of group convolutions \cite{xie2017aggregated}. It utilizes the depthwise separable convolutions \cite{sifre2014rigid}, which are the core building blocks of many efficient networks, to structure the lightweight network.
MobileNetV2, which is based on MobileNetV1, is a more efficient and effective lightweight network.
In addition to depthwise separable convolutions, the novel inverted residual with linear bottleneck is proposed. MobileNetV2 achieves the state-of-the-art speed-accuracy tradeoff among many lightweight networks. In this paper, we develop a real-time high-performance semantic segmentation method based on MobileNetV2.
%Several methods have also been developed to change the internal connection structure inside the convolution blocks, such as ShuffleNet.

\section{Preliminary Knowledge}
\label{preliminary}
In this section, we mainly present the preliminary knowledge of atrous convolution and atrous spatial pyramid pooling, which constitute the important basis of our method.

%This has also been observed independently by [81].3

Atrous convolution has a long history in the filed of signal processing. It is initially created for efficient computation of undecimated wavelet transform in the ``algorithme \`a trous'' scheme \cite{holschneider1990real}. Atrous convolution is a term that was firstly used in \cite{papandreou2015modeling} and introduced into semantic segmentation in DeepLab \cite{chen2014semantic}. The same operation was later called dilated convolution (motivated by the fact that such an operation corresponds to the standard convolution with a dilated filter) in \cite{yu2015multi}.

 The idea of atrous convolution is to convolve the input image with the upsampling filter produced by inserting zeros (or holes) between each pair of consecutive non-zero filter values. %See Figure \ref{fig:atrous_conv} for an illustration.
Compared with the standard convolution, atrous convolution can effectively increase the size of receptive fields and generate denser feature maps without increasing the amount of computation and the number of parameters.
%Here, we utilize the mathematical model to understand atrous convolution.% For the sake of simplicity, we only show one-dimensional (1-D) case here.

\begin{figure}[!t]
\centerline{
\includegraphics[scale=0.3]{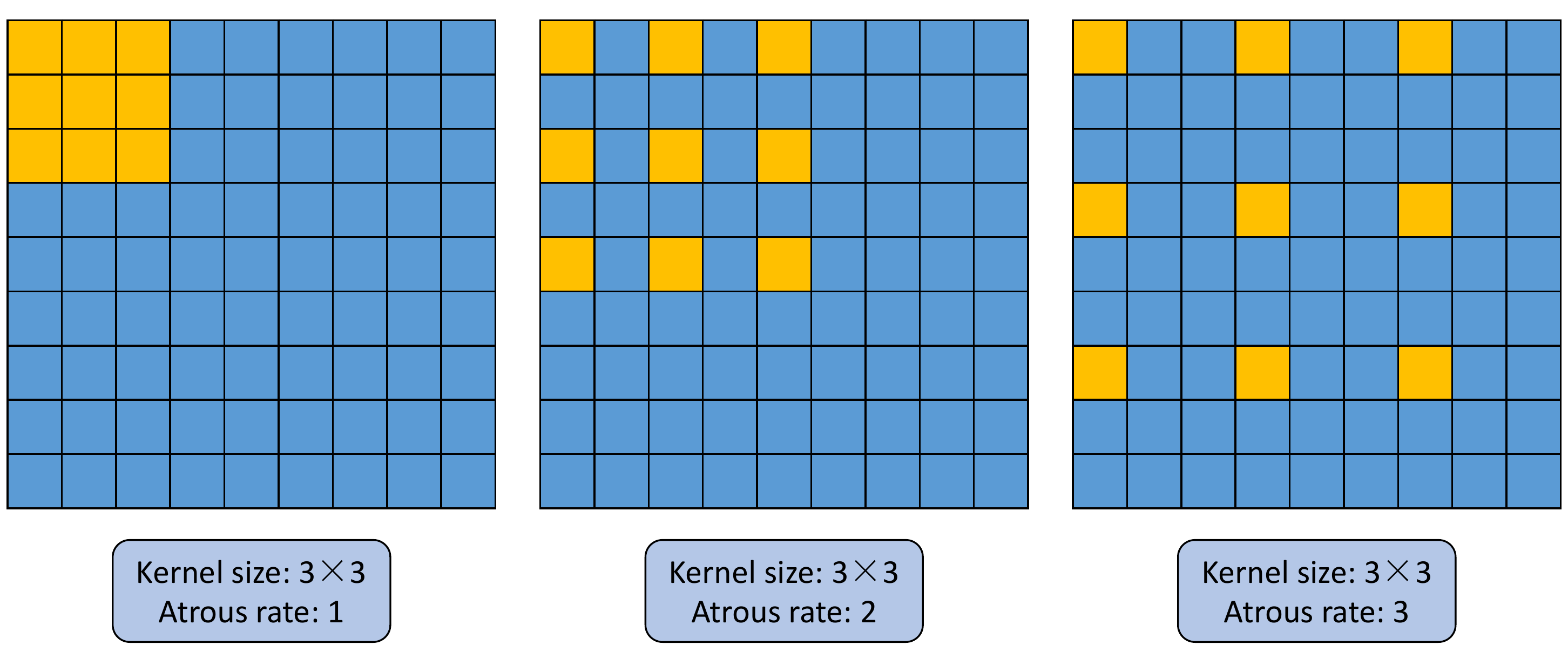}}
\caption{Illustration of atrous convolution in 2D with different atrous rates (1, 2 and 3), where kernel size = $3\times3$.}
\label{fig:atrous_conv}
\end{figure}

%Let $P_{M\times N,d}(x)$  represent the atrous convolution layer of each branch, where
Mathematically, let $\textbf{X}[i,j]$ and $P_{k,d}(\textbf{X})[i,j]$ respectively denote the input and output signals of atrous convolution at location $(i,j)$, where $k\times k$ and $d$ respectively denote the size of the convolution kernel and the atrous rate (indicating the number of zeros to be inserted).
 An atrous convolution $(2D)$ can be written as follows:
\begin{equation}
\label{eqn:lcenter}
P_{k,d}(\textbf{X})[i,j]=\sum_{m=1}^{k}\sum_{n=1}^{k}\textbf{X}[i+d \cdot m, j+d \cdot n] \cdot \textbf{W}[m,n]
\end{equation}
%where $d$ is the atrous rate which indicates the number of zeros to be inserted, and
where $\textbf{W}$ denotes a convolution kernel with the size of $k \times k$. In particular, if the atrous rate is 1 (i.e., $d = 1$), the atrous convolution becomes the standard convolution. See Fig.~\ref{fig:atrous_conv} for an illustration.

\begin{figure}[!t]
\centerline{
\includegraphics[scale=0.48]{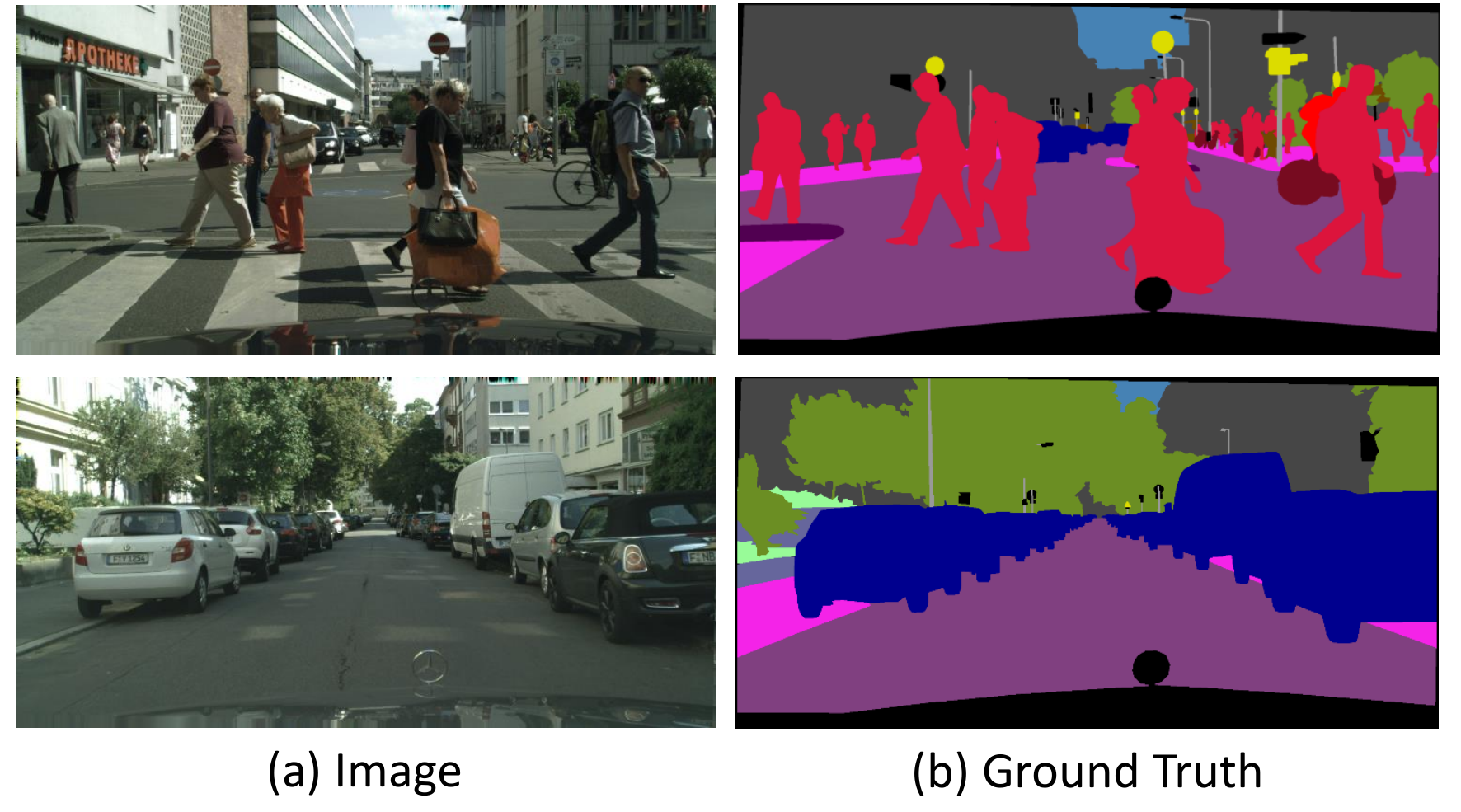}}
\caption{Illustration of the challenging multi-scale variations on the Cityscapes dataset. From the images, the same category of objects (such as humans or cars) varies largely in scale.}
\label{fig:different_distance}
\end{figure}

One of the challenges for DCNN-based semantic segmentation is caused by the existence of objects at multiple scales. The sizes of objects are usually different and frequently change (even for the same category of objects), specifically for the segmentation of urban street scenes (see Fig.~\ref{fig:different_distance} for an illustration). The same category of objects (such as humans or cars) in a scene shows different sizes due to their different distances to the camera.
Hence, how to correctly extract and encode the multi-scale information is a great challenge.

There are two common approaches to deal with the multi-scale problem. A standard approach is to feed the different sizes of an image into a network and then aggregate the feature maps or score maps together \cite{chen2016attention}. This is helpful to handle the multi-scale objects and improve the segmentation accuracy. However, it usually requires a large amount of computation.

Another approach is to take advantage of computationally efficient Atrous Spatial Pyramid Pooling (ASPP), which is based on spatial pyramid pooling \cite{lazebnik2006beyond}.
ASPP uses parallel atrous convolution branches with different atrous rates (corresponding to different sizes of receptive fields) to generate the feature maps covering multiple scales of receptive fields. Therefore, ASPP is able to effectively encode the multi-scale information and thus boost the final segmentation performance.

%The objective of Therefore, it concatenates the feature maps generated by the parallel atrous convolutions with multiple atrous rates .

\begin{figure}[!t]
\centerline{
\includegraphics[scale=0.55]{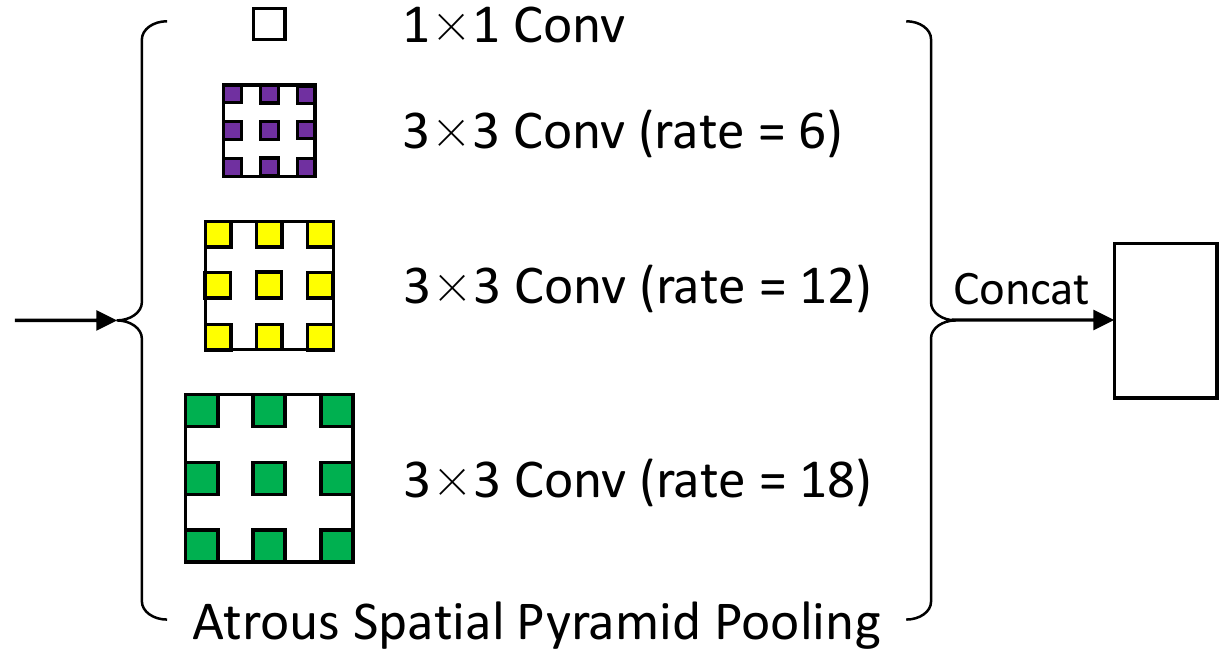}}
\caption{Illustration of Atrous Spatial Pyramid Pooling. ASPP utilizes multiple parallel branches with different atrous rates to capture multi-scale contexts.}
\label{fig:previous_ASPP}
\end{figure}

%As shown in Fig.~\ref{fig:previous_ASPP}, the whole ASPP consists of one $1\times1$ convolution, three $3\times3$ convolutions with atrous rates = (6, 12, 18) and the image-level pooling \cite{liu2015parsenet}.
An example of ASPP, which consists of one $1\times1$ convolution, and three $3\times3$ convolutions with atrous rates being equal to (6, 12, 18), is given in Fig.~\ref{fig:previous_ASPP}. The four different atrous convolution branches (corresponding to 4 different atrous rates) have the same input and their outputs are combined. The final output can be viewed as the sampling of all feature maps from the four branches with different scales of receptive fields. Therefore, ASPP is able to robustly segment objects at multiple scales effectively.

Similarly, let $P_{k,d}(\textbf{X})$ represent the atrous convolution layer of each branch, where $k \times k$ and $d$ denote the size of each atrous convolution kernel and the corresponding atrous rate, respectively. Therefore, ASPP can be written as follows:
\begin{equation}
\label{eqn:lpos}
\textbf{Y}=P_{1, 1}(\textbf{X}) \odot P_{3, 6}(\textbf{X}) \odot P_{3, 12}(\textbf{X}) \odot P_{3, 18}(\textbf{X})
\end{equation}
where $\textbf{X}$ and $\textbf{Y}$ are the input and output of ASPP, respectively.
Note that `$\odot$' represents the concatenation operation of feature maps generated by atrous convolution with different arous rates.

%For the rest of this section, we first illustrate the entire network framework and processing flow, and then we detailedly explain the principle of each component in turn.

\section{The Proposed Method}
\label{proposedmethod}

\begin{figure*}[!t]
\centerline{
\includegraphics[width=7in]{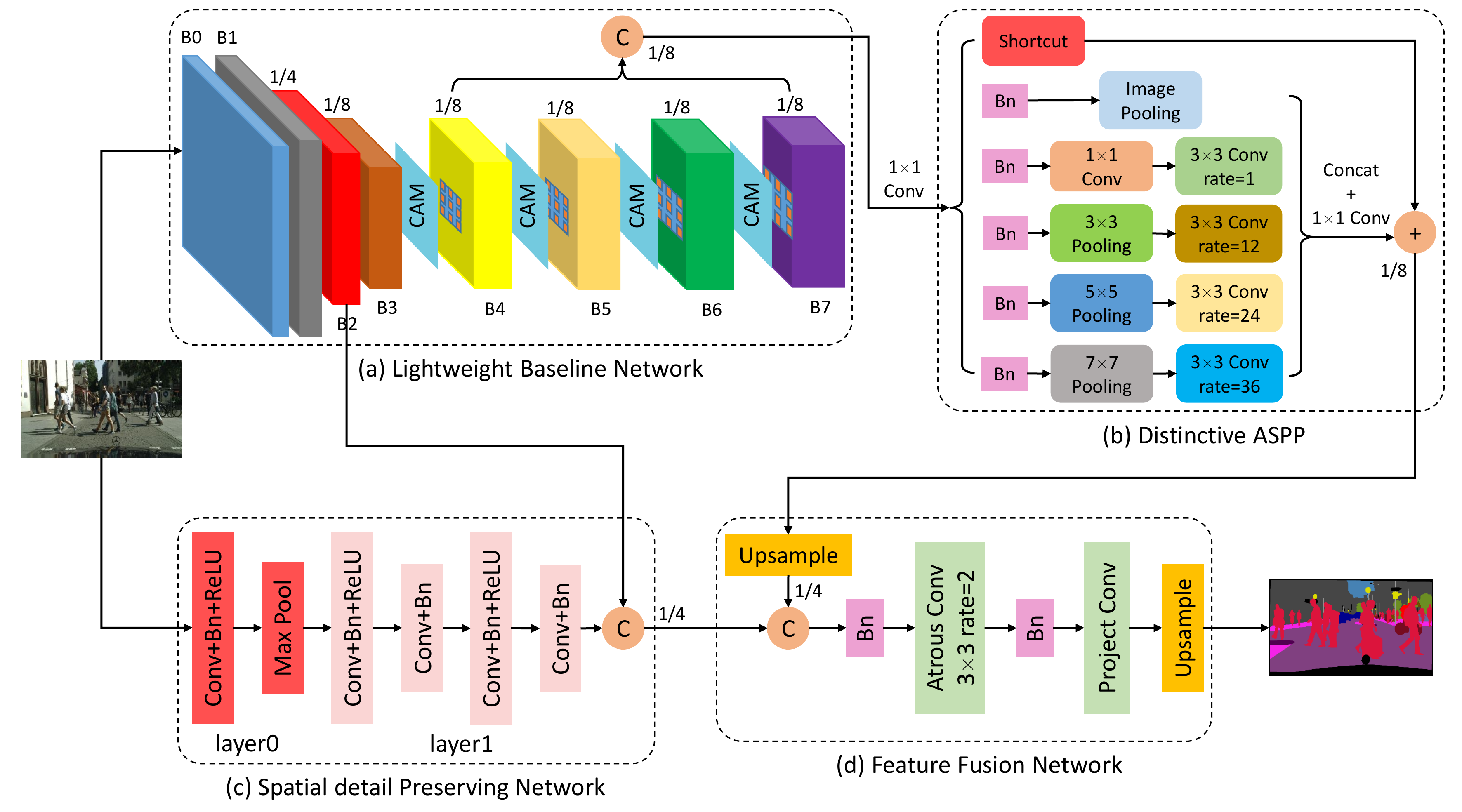}}
\caption{The overall framework of the proposed real-time high-performance semantic segmentation method. (a) is the proposed Lightweight Baseline Network with Atrous Convolution and Attention (LBN-AA).
%Here, B0-B7 denotes the Block0-Block7.
(b) is the improved Distinctive Atrous Spatial Pyramid Pooling (DASPP). (c) is the Spatial detail-Preserving Network (SPN). (d) is the  Feature Fusion Network (FFN).}
\label{fig:network_framework}
\end{figure*}

In this section, we introduce the proposed real-time high-performance semantic segmentation method in detail. An overview of the proposed method is introduced in Section \ref{overview}. Each component of the proposed method is described from Section \ref{lbnaa} to Section \ref{ffn}.
%
%our efficient architecture for
%%real-time semantic segmentation.
%In this section, an overview of the proposed MS-CFB algorithm
%for face recognition is introduced in Section 3.1. The detailed design process of a CFB and feature extraction based on CFBs are
%described in Sections 3.2 and 3.3, respectively. Classification rule is
%presented in Section 3.4. The complete algorithm is given in
%Section 3.5. We discuss the proposed algorithm in Section 3.6.

\subsection{Overview}
\label{overview}
The proposed real-time high-performance semantic segmentation method consists of four main components: the Lightweight Baseline Network with Atrous Convolution and Attention (LBN-AA), the Distinctive ASPP (DASPP), the Spatial detail-Preserving Network (SPN) and the Feature Fusion Network (FFN).
An overview of the whole network is illustrated in Fig.~\ref{fig:network_framework}.

%Firstly, we downsample the input images to a moderate size.
%%(e.g., we resize the input images from the size of $1024\times2048$ to $448\times896$ for training and $400\times800$ for testing).
%Image resolution is one of the critical factors that affect the speed.
%The semantic segmentation methods consume more computational time, when the resolution of an input image is larger. Therefore, limiting the resolution of the input image can reduce the complexity of our network and improve the inference speed.
Firstly, the input images are fed into LBN-AA. Note that many real-time semantic segmentation methods (e.g., \cite{badrinarayanan2015segnet, zhao2017icnet}) downsample the feature maps to 1/32 or 1/16 of the original input image size, which leads to poor accuracy. In contrast, LBN-AA is able to generate the larger size of feature maps which are 1/8 of the original input image size due to the usage of atrous convolution. Thus, LBN-AA will not lose too much information. Secondly, the feature maps extracted from LBN-AA are directly fed into DASPP to segment the objects at multiple scales and extract the high-level semantic information.
Meanwhile, the input images are also fed into SPN to further extract sufficient spatial information.
The semantic path (LBN-AA+DASPP) and the spatial path (SPN) are simultaneously computed, which considerably improves the efficiency.
Thirdly, the outputs from the two paths are effectively combined by FFN to obtain the predicted results.
Finally, we utilize the simple bilinear interpolation \cite{chen2018deeplab, liu2015semantic} to upsample the predicted results to the original image size. %of $1024\times2048$.

In the following, we will introduce the four components of the proposed method in detail.

\subsection{Lightweight Baseline Network with Atrous Convolution and Attention (LBN-AA)}
\label{lbnaa}

Our proposed LBN-AA (shown in Fig.~\ref{fig:network_framework}(a)) is based on MobileNetV2 \cite{sandler2018inverted} pretrained on the ImageNet dataset \cite{russakovsky2015imagenet}. MobileNetV2 is the latest lightweight image classification network, with the characteristics of fast speed, little computational time and small memory consumption.
%Note that MobileNetV2 is mainly developed for object classification (the output is a probability vector).  Moreover, MobileNetV2 substantially reduces the resolution of feature maps in the last bottleneck layer (the size of the output is only $1/32$ of the input image size).
Note that MobileNetV2 is mainly developed for object classification, which substantially reduces the resolution of the input (the size of the output in the last bottleneck layer is only $1/32$ of the original input image size and the final output is a probability vector).
 However, for semantic segmentation, we need to ensure that the size of output feature maps is moderately large (otherwise the spatial details will significantly lose).
Hence, we appropriately modify the original structure of MobileNetV2 to serve as our baseline network.

Generally, we remove all the convolution and pooling layers after the last bottleneck layer of MobileNetV2 and then obtain a simplified version of MobileNetV2, which contains the initial convolution layer with 32 channels and 19 residual bottlenecks followed.
The output feature maps of this simplified MobileNetV2 are only 1/32 of the original input image size.
In other words, a significant amount of spatial details will lose. It is intuitive to reduce the depth of the network to obtain large feature maps. However, the discriminability of feature maps may substantially reduce, since the size of receptive fields becomes small. Therefore, we propose to effectively combine the simplified MobileNetV2 with atrous convolution. Atrous convolution can be used to obtain the larger size of receptive fields and denser feature maps than standard convolution.
Although the introduction of atrous convolution increases the computational time, combining the simplified MobileNetV2 and atrous convolution can still achieve a great balance between accuracy and speed (since the simplified MobileNetV2 is an extremely lightweight network).

In addition, the importance of each channel of a feature map is often different.
 As a matter of fact, the convolutional filter in DCNN plays the role of pattern detector, and each channel of a feature map can be viewed as a response activation of the corresponding filter. Some channels encode more spatial and context information than other channels for semantic segmentation. In particular, some irrelevant information (such as background clutter or noise) in some channels might be improperly learned when training the deep neural network. The irrelevant information can greatly affect the extraction of important semantic information, thus leading to the performance decrease of semantic segmentation. This problem becomes more serious for lightweight networks, such as MobileNetV2.

To solve the above problem, inspired by the Squeeze-and-Excitation (SE) block \cite{hu2018squeeze}, we further append several Convolutional Attention Modules (CAM)  in the modified MobileNetV2 to select the informative channels. Thus, we take advantage of the weights generated by CAM to guide the network learning and thus the weighted feature maps can be obtained. Such an approach is beneficial for emphasizing the important information and suppressing the irrelevant information.

\begin{figure}[!t]
\centerline{
\includegraphics[scale=0.45]{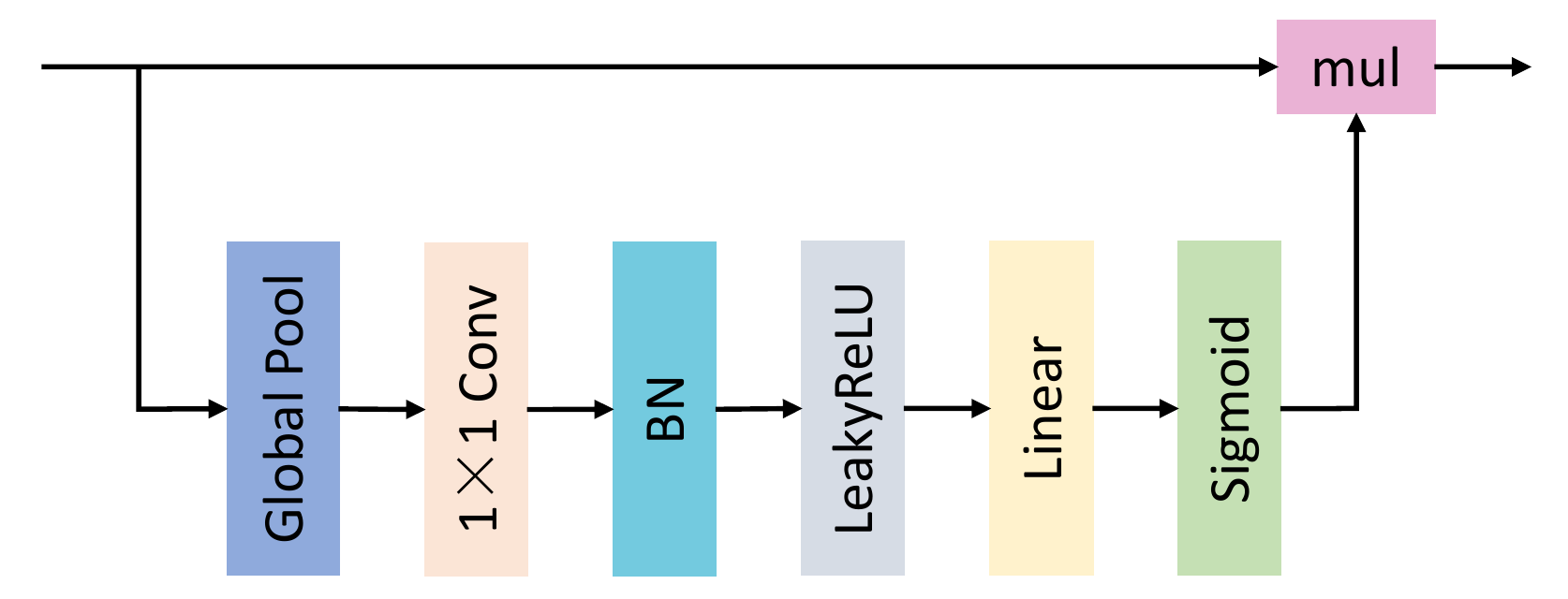}}
\caption{Network architecture of CAM.}
\label{fig:attention}
\end{figure}

The detailed network configuration of LBN-AA is given in Table \ref{tab:modified_mobilenetv2}. More specifically, we modify the simplified MobileNetV2 as follows. Firstly, the frontal layers of MobileNetV2 (i.e., block0 to block3 in Table \ref{tab:modified_mobilenetv2}) remain unchanged until the resolution of feature maps drops to 1/8 of the original input image size.
Secondly, from block4 to block7, we add atrous convolution to the first bottleneck layer of each subsequent block and set the stride to 1 so that the resolution of feature maps keeps unchanged. Motivated by the DeepLabv2 method \cite{chen2018deeplab}, which employs the different hierarchies of atrous convolutions, we adopt different atrous rates (i.e., 2, 4, 8 and 16 for block4 to block7, respectively)  \cite{terzopoulos1986image} in LBN-AA.
 The proposed CAMs are fed into the block4, block5, block6 and block7 in sequence to select the important information for segmentation.
The network architecture of CAM is shown in Fig.~\ref{fig:attention}.
CAM firstly employs the global average pooling and $1\times 1$ convolution followed by Batch Normalization (BN) \cite{ioffe2015batch} and LeakyReLU \cite{maas2013rectifier} to encode the importance of output features into a vector.
Note that the number of input channels is reduced by the $1\times 1$ convolution operation, which effectively improves the efficiency of CAM.
Then, CAM utilizes the linear operation (i.e., the fully-connected layer) and the Sigmoid function to obtain the attention vector.
Finally, the different channels of the feature maps are weighted according to the attention vector.

%Convolutional Neural Networks with Alternately Updated Clique
%Our convolutional attention modules are placed in front of the block4, block5, block6 and block7 in sequence to refine the learning process.}

%The module first contains a global average pooling and a convolution followed Batch Normalization (BN) \cite{ioffe2015batch} and LeakyReLU \cite{maas2013rectifier} to encode the importance of output features into a vector. Note that the number of input channels will be reduced in convolution.
%a quarter of the original number in convolution.

\begin{table}[!t]
\renewcommand{\arraystretch}{1.3}
\caption{The Detailed Network Configuration of Our LBN-AA. Each Operation is Repeated $n$ Times. $t$ Denotes the Expansion Factor of Bottleneck. The Strides of the First Layer and All Others are Set to $s$ and 1, Respectively. The Parameter $d$ Denotes the Atrous Rate of the First Layer for Each Block. Note That the Stride $s$ is Set to 1 When the Atrous Rate is Not Equal to 1.}
\label{tab:modified_mobilenetv2}
\centering
\scalebox{0.91}{
\begin{tabular}{c|c|c|ccccc} %|c||c|
\Xhline{1.2pt}
{Block} & {Input} & {Operation} & {$t$} & {$c$} &	{$n$} & {$s$} & {$d$} \\
%\Xhline{1.2pt}
\hline
\hline
\textbf{block0} & $448\times896\times3$ & conv2d & - & 32 & 1 & 2 & 1 \\
\hline
\textbf{block1} & $224\times448\times32$ & bottleneck & 1 & 16 & 1 & 1 & 1 \\
\hline
\textbf{block2} & $224\times448\times16$ & bottleneck & 6 & 24	& 2 & 2 & 1 \\
\hline
\textbf{block3} & $112\times224\times24$ & bottleneck & 6 & 32 & 3 & 2 & 1 \\
\hline
\textbf{block4} & $56\times112\times32$ & bottleneck & 6 & 64 & 4 & 1 & 2 \\
\hline
\textbf{block5} & $56\times112\times64$ & bottleneck & 6 & 96 & 3 & 1 & 4 \\
\hline
\textbf{block6} & $56\times112\times96$ & bottleneck & 6 & 160 & 3 & 1 & 8 \\
\hline
\textbf{block7} & $56\times112\times160$ & bottleneck & 6 & 320 & 1 & 1 & 16 \\
\Xhline{1.2pt}
\end{tabular}
}
\end{table}

%\textcolor{red}{%Skip connection is widely used in semantic segmentation networks, specifically for the encoder-decoder structure. }The main idea of skip connection is to combine the feature maps from the higher layers and those from the lower layers (with larger resolutions), thus improving the segmentation results.}

\begin{figure}[!t]
\centerline{
\includegraphics[scale=0.5]{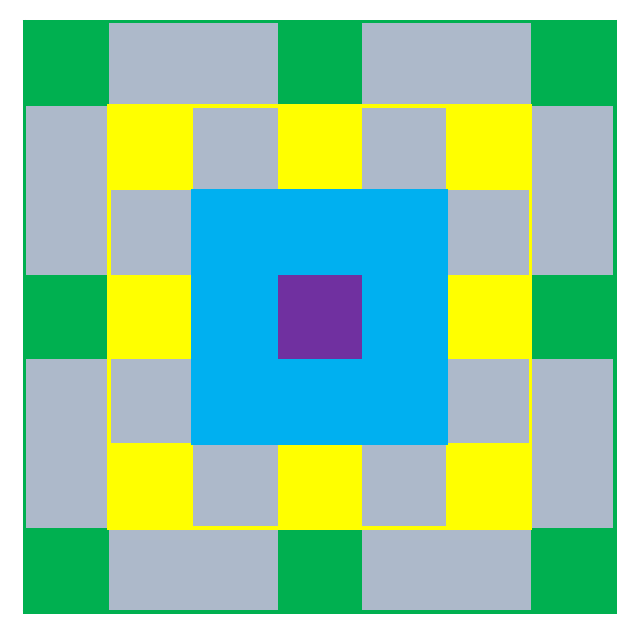}}
\caption{Stacking the atrous convolution layers with different atrous rates generates a denser sampling to alleviate the gridding issue.}
\label{fig:grid_issue}
\end{figure}

From Table \ref{tab:modified_mobilenetv2}, we can see that the sizes of the output feature maps from block3 to block7 are 1/8 of the original input image size. Therefore, we further employ the dense skip connections by concatenating the feature maps together (from block4 to block7) instead of using element-wise addition as the outputs of LBN-AA. There are two benefits of using the dense skip connections. On one hand, more information can be utilized for training the network, since we concatenate information from multiple layers. Therefore, the segmentation accuracy can be greatly improved without increasing much computational burden.
On the other hand, the gridding issue caused by the atrous convolution operations \cite{wang2018understanding, mehta2018espnet} can be alleviated to some extent (see Fig.~\ref{fig:grid_issue} for an illustration).
According to the preliminary knowledge mentioned above, zeros are padded between two consecutive non-zero values in an atrous filter. Because only the locations of non-zero values are sampled, the receptive fields cover an area with the checkerboard pattern, which leads to the loss of some neighboring information (see the gray grids in Fig.~\ref{fig:grid_issue}).
%\textcolor{red}{%This problem is more evident when the atrous rate is larger.}
Note that the blocks from block4 to block7 have different atrous rates. Therefore, LBN-AA is able to generate a dense sampling by concatenating the feature maps from these blocks, where the neighboring information from different blocks can complement each other \cite{wang2018understanding}.% Such a manner effectively alleviates the gridding issue.

{It is worth noting that BiSeNet \cite{yu2018bisenet} also employs an attention module called Attention Refinement Module (ARM) for semantic segmentation. However, CAM and ARM are significantly different. Firstly, the motivations of CAM and ARM are different. In this paper, we combine CAM and the convolution block in the lightweight network to effectively guide the network learning. In contrast, ARM is mainly used to refine the features from the outputs of the context path. Secondly, the network architectures of CAM and ARM are different. LBN-AA is based on the lightweight network (MobileNetV2), where the information flow in the network is restricted. The use of  LeakyReLU and linear operation in CAM is beneficial for back-propagating information during the network training (note that LeakyReLU allows for a small, non-zero gradient even when the unit is saturated and not active \cite{maas2013rectifier}), and extracting informative features. In contrast, BiSeNet is based on the Xception model (which is more complex than MobileNetV2) and the use of ReLU in ARM is sufficient for network training.}

\subsection{Distinctive Atrous Spatial Pyramid Pooling (DASPP)}

%The objective of ASPP \cite{chen2018deeplab} is to effectively encode multi-scale information and thus boost the final segmentation performance. Therefore, it concatenates the feature maps generated by the parallel atrous convolutions with multiple atrous rates (corresponding to different sizes of receptive fields).

According to the theory of Receptive Fields (RFs) in human visual systems \cite{liu2017receptive, wandell2015computational}, the diverse inputs are beneficial to extract distinctive features. However, from Fig.~\ref{fig:previous_ASPP}, each branch of ASPP employs the same input feature maps (the same input is used for each branch). Therefore, the output feature maps tend to be less distinctive, thus leading to confusion between objects and context. Moreover, for each pixel in the output feature maps, only 9 distant pixels (if a $3\times3$ kernel is used) in an input are used during each atrous convolution operation. Therefore, the pixels of an input feature map are not fully and effectively exploited.

In view of the shortcomings of previous ASPP, we propose the distinctive ASPP (DASPP), see Fig.~\ref{fig:network_framework}(b). More specifically, we add the average pooling layers with different sizes before the atrous convolution operation of each branch. In this paper, we adopt the pooling sizes of $3\times3$, $5\times5$ and $7\times7$ in the parallel atrous convolution branches, respectively. The corresponding atrous rates of $3\times3$ atrous convolution operation in the three branches are 12, 24 and 36, respectively. Moreover, we change the original $1\times1$ convolution branch in ASPP to $1\times1$ convolution and $3\times3$ convolution operations, which effectively improve the capacity of feature extraction.
The image-level pooling layer \cite{liu2015parsenet} for capturing the global context information is still adopted.

It is worth pointing out that the atrous rate in atrous convolution should be properly chosen to ensure the performance of semantic segmentation. In particular, the atrous rate in atrous convolution cannot be set to a large value if the size of input images is small. This is mainly
because that some pixels in the input image are not exploited if a large value of  atrous rate is used
(note that the atrous convolution is sparse convolution). Moreover, the spatial information is lost since the receptive field is increased. In this paper, the atrous rates of atrous convolution in the three branches are set to be the same as those used in DeepLabv2 \cite{chen2018deeplab}.

\begin{figure}[!t]
\centerline{
\includegraphics[scale=0.4]{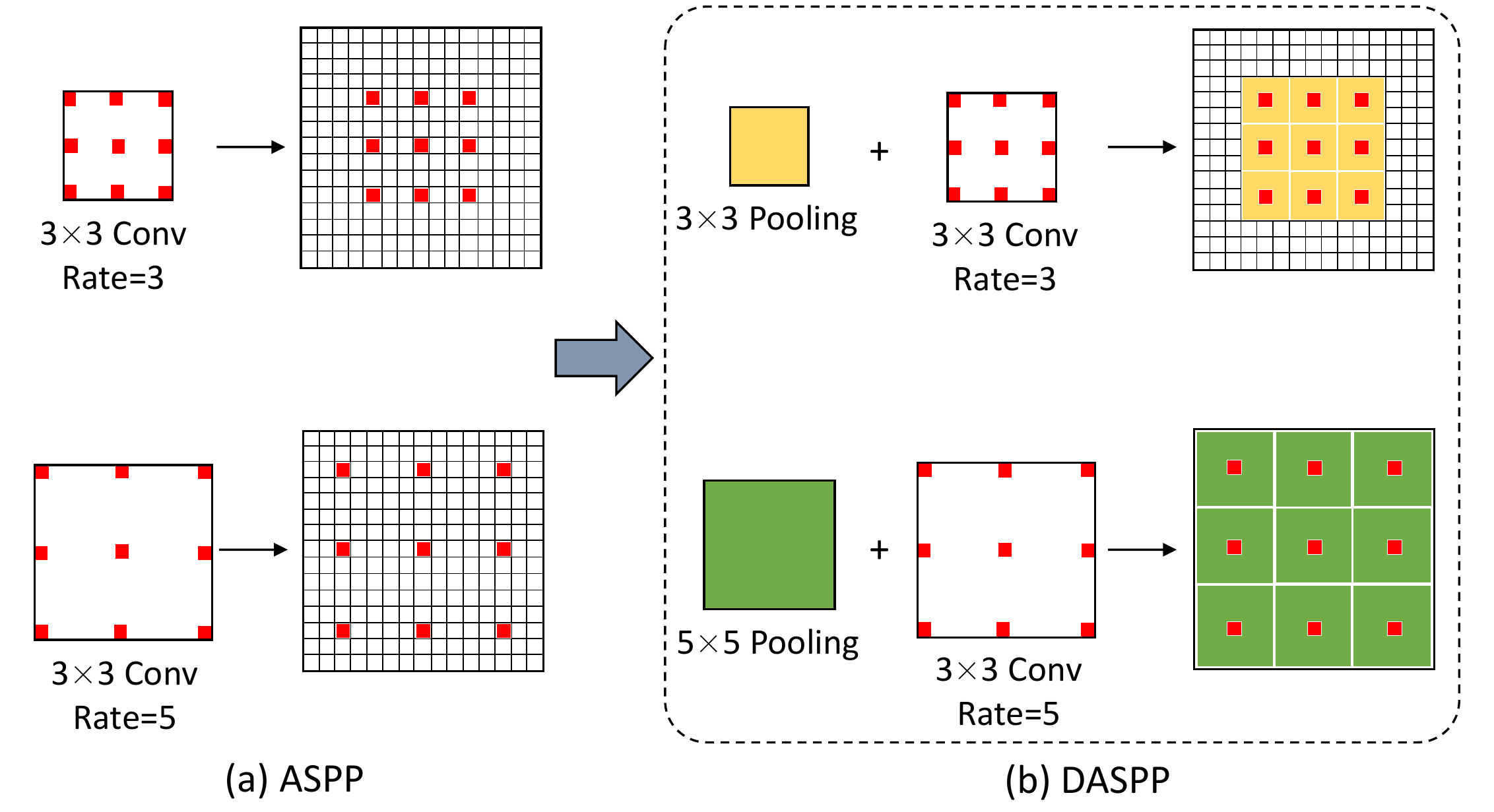}}
\caption{Comparison between (a) ASPP and (b) DASPP.
%In view of the shortcomings of previous ASPP, our DASPP utilizes the simple but effective average pooling operations with different sizes.
Here, we only use a simple example to illustrate the principle.}
\label{fig:improved_ASPP}
\end{figure}

Mathematically, DASPP can be formulated as follows:
\begin{equation}
\label{eqn:lpos}
\begin{split}
\textbf{Z}=&I _{pooling}(\textbf{X}) \odot P_{3, 12}(H_{pooling,3}(\textbf{X})) \odot P_{3, 24}(H _{pooling,5}(\textbf{X}))\\
& \odot P_{3, 36}(H_{pooling,7}(\textbf{X})) \odot P_{3, 1}(P_{1,1}(\textbf{X}))
\end{split}
\end{equation}
where $\textbf{Z}$ represents the output of DASPP. $I_{pooling}(\textbf{X})$ denotes the image-level average pooling.
$H_{pooling,k}(\textbf{X})$ represents the average pooling operation with the pooling size of $k\times k$. Note that we employ one $1\times1$ filter with 128 channels to reduce the input dimension before DASPP. After the concatenation of the feature maps from all the branches, these feature maps are passed through one $1\times1$ filter to reduce the amount of channels to 128 for efficient computation.

Compared with ASPP, the proposed DASPP takes advantage of the different sizes of pooling operations to extract more distinctive features and exploit the neighboring information of each pixel, as shown in Fig.~\ref{fig:improved_ASPP}.
 In this way, the inputs for each atrous convolution operation in the parallel branches are different, while the neighboring pixels of an input are fully used (see Fig.~\ref{fig:improved_ASPP}(b)). Therefore, DASPP can obtain more distinctive and informative feature maps than ASPP.

Each branch in the proposed DASPP only includes 128 channels, where BN is also used for each branch to accelerate the network training. Moreover, considering the merits of shortcut \cite{he2016deep}, we also apply the shortcut to reuse the input feature maps of DASPP and facilitate the information flow. Since the padding operations are used, the output resolution of each branch is equal to the input one. All the feature maps are finally fused by using the simple element-wise addition.

Recently, Xie et al. \cite{xie2018vortex} propose the vortex pooling,
which also adds the  average pooling layers before the parallel atrous convolution layers
 for semantic segmentation. The vortex pooling bears some similarities to the proposed DASPP. However, DASPP and the vortex pooling are different in the following aspects.

{Firstly, the motivations of vortex pooling and DASPP are different. The proposed DASPP is motivated by the observation that the diverse inputs play an important role in extracting distinctive features \cite{liu2017receptive}. Hence, DASPP takes advantage of the different sizes of pooling operations to generate the diverse inputs.
In contrast, the vortex pooling is motivated by the assumption that the descriptors near and far from the target pixel respectively contain the related semantic information
and provide the contextual information. Therefore, it uses the average pooling with different kernels to get fine and coarse representations.
Secondly, the specific network details, configuration and hyper-parameters are also different.
For example, the atrous rates and pooling sizes used in DASPP are different from those used in the vortex pooling, and the standard convolution branch is also different.
DASPP concatenates the outputs from the four atrous convolution branches, and then calculates the element-wise addition of the shortcut connection and the concatenated outputs. In contrast, the vortex pooling directly performs the element-wise addition of the outputs from different atrous convolution branches.
Finally, DASPP (the semantic branch) is tightly combined with SPN (the spatial branch) by using FFN (the fusion network) to obtain the excellent semantic segmentation performance. Unlike DASPP, the vortex pooling is combined with the global average pooling to obtain the context features in \cite{xie2018vortex}.}

%Our specific modifications are as follows: (a) We retain the original $3\times3$ atrous convolution of each parallel branch, and then add an average pooling layer with different pooling sizes in the front of atrous convolutions to obtain more information and increase the diversity of features.  (b) We change the original $1\times1$ convolution branch to $1\times1$ and $3\times3$ convolutions, which improves the capacity of feature extraction. (c) We still . Therefore, our improved ASPP consists of (a) one $1\times1$ convolution followed by one $3\times3$ convolution, (b) three average pooling operations with sizes = ($3\times3$, $5\times5$, $7\times7$) followed by corresponding three $3\times3$ atrous convolutions with rates = (12, 24, 36), and (c) the global average pooling layer, as shown in Figure \ref{fig:network_framework}(b).

%Simply speaking, the functions of ASPP are to  eventually. The ASPP we want to improve here is from \cite{chen2017rethinking}, which consists of one $1\times1$ convolution, three $3\times3$ convolutions with atrous rates = (6, 12, 18) and the image-level pooling \cite{liu2015parsenet}, as shown in Figure \ref{fig:previous_ASPP}. Different from \cite{chen2018deeplab}, this ASPP incorporates batch normalization \cite{ioffe2015batch} and global context information to the model. Although it is undeniable that this ASPP can effectively capture multi-scale information and boost performance, it still has some problems, and there is room for improvement.

\subsection{Spatial detail-Preserving Network (SPN)}
A high-performance semantic segmentation network requires both sufficient semantic information with sizeable receptive fields and rich spatial information with fine details.
Generally speaking, the deeper a network is, the larger the receptive fields are, and the more semantic information is extracted. However, with the increase of network depth, the spatial information is seriously lost. Many modern methods usually compromise the spatial information to obtain larger receptive fields. In other words, they ignore that the spatial information is also important for predicting a fine result for the semantic segmentation task.

In our proposed method, the outputs of DASPP mainly encode the rich semantic information while the spatial information is only partially preserved. Therefore, we design a Spatial detail-Preserving Network (SPN), which is able to preserve rich spatial details and generate high-resolution feature maps to refine the output results of DASPP. The structure of SPN is shown in Fig.~\ref{fig:network_framework}(c).

%However, in the semantic segmentation task,
%
%We all know that the deeper a network is, the larger the receptive fields in the upper parts of network will be and the more semantic information will be acquired, but the spatial information in the lower parts of network will be constantly lost. Many modern approaches usually compromise the spatial information to obtain larger receptive fields. However, they ignore that spatial information is equally important for predicting a detailed result in the semantic segmentation tasks.

%So far, the network we discussed above can already achieve a good segmentation effect and contains sufficient semantic information. Although the network also contains some spatial information due to atrous convolution, it is still far from enough. This prevents the performance of our network from achieving a higher level. Different from previous approaches which only consider semantic or spatial information, we aim to better combine these two seemingly contradictory aspects.

To avoid excessive computation, SPN is designed to be a shallow neural network. Specifically, we advocate a compact version of ResNet-18 (which is pretrained on the ImageNet dataset) and adopt the first two layers (i.e., layer0 and layer1) of ResNet-18 as SPN (see Table \ref{tab:resnet18} for more detail).  In addition, we also concatenate
the outputs from layer1 and the block2 in LBN-AA as the
final output of SPN.
 The final output feature maps of SPN are 1/4 of the original input image size and include 88 channels. Although the resolution of feature maps obtained by SPN is relatively large, the computational cost is small (since only shallow convolution layers are used from the compact version of ResNet-18).

Note that some methods \cite{yu2018bisenet} propose to use several simple convolution layers with a small number of filters to preserve spatial information.
For example, the spatial path of BiSeNet \cite{yu2018bisenet} contains three convolution layers, followed by
BN and ReLU. However, previous works \cite{liu2017receptive}  have shown that training the network from the scratch is not a trivial task.
In this paper, the proposed SPN leverages the strategy of combining the pretrained backbones and fine-tuning, which is beneficial for improving the final performance. Since the initial weights of SPN (pretrained from the large dataset) are useful for extracting
the low-level features, the training process can be easily converged.
Furthermore, the trained SPN model can obtain the improved accuracy.

\begin{table}[!t]
\renewcommand{\arraystretch}{1.3}
\caption{Network Architecture of SPN.}
\label{tab:resnet18}
\centering
\scalebox{0.9}{
\begin{tabular}{c|c|c|ccc}
\Xhline{1.2pt}
{Layer} & {Input size} & {Operation} & {Kernel} & {Channel} & {Stride} \\
\hline
\hline
\multirow{2}{*}{layer0} & $448\times896$ & conv2d & $7\times7$ & 64 & 2 \\
\cline{2-6} & $224\times448$ & max pool & $3\times3$ & - & 2 \\
\hline
\multirow{4}{*}{layer1} & $112\times224$ & conv2d & $3\times3$ & 64	& 1 \\
\cline{2-6} & $112\times224$ & conv2d & $3\times3$ & 64 & 1 \\
\cline{2-6} & $112\times224$ & conv2d & $3\times3$ & 64 & 1 \\
\cline{2-6} & $112\times224$ & conv2d & $3\times3$ & 64 & 1 \\
\Xhline{1.2pt}
\end{tabular}
}
\end{table}

\subsection{Feature Fusion Network (FFN)}
\label{ffn}
The feature maps generated from two branches (LBN-AA+DASPP and SPN) effectively encode different types of information. That is, the feature maps (deep features) generated from LBN-AA and DASPP mainly encode sufficient high-level semantic information, while the spatial details (shallow features) captured by SPN mainly provide rich low-level spatial information. Thus, we need to fuse the deep and shallow features. Since the features from the two branches belong to different levels of feature representation, using simple fusion techniques (such as element-wise addition or concatenation along the channels) may decrease the accuracy. Therefore, we propose a simple but practical Feature Fusion Network (FFN) to effectively fuse these features, as shown in Fig.~\ref{fig:network_framework}(d).

Specifically, we firstly concatenate the feature maps from the two branches along the channels. Then, we apply the Batch Normalization operation to shorten the feature distances and balance the feature scales. Note that the pixels at the same position of two feature maps are not necessarily similar and they may be similar to the adjacent pixels at the position. Therefore, we apply an atrous convolution with the size of $3\times3$ and the arous rate $d = 2$ to fuse the features. This operation is able to effectively combine the feature information from adjacent pixels around the target pixel instead of depending on just one position.
After that, a projection convolution with the size of $1\times1$ is used to reduce the number of output channels to the number of semantic categories (from 216 to 19). A BN layer is also used between the atrous convolution and projection convolution. Finally, we utilize the simple and efficient bilinear interpolation, which only requires a few parameters and can achieve similar accuracy as the transposed convolution \cite{zeiler2011adaptive}, to directly upsample the fused results to the original input image size.

We should point out that ICNet \cite{zhao2017icnet} employs the Cascade Feature
Fusion (CFF) module to combine the cascade features from the different resolution inputs. CFF respectively performs the atrous convolution with BN and the projection convolution with BN for two inputs and then concatenates the outputs from the two convolutions. In contrast, FFN concatenates the inputs from different branches, and then performs the atrous convolution and projection convolution in a cascaded way. This is mainly because that the number of channels for two inputs from the semantic branch and the spatial branch is different. Therefore, we cannot
directly use the CFF module in FFN.

\section{Experimental Evaluation}
\label{experiment}
In this section, we conduct extensive experiments to evaluate the effectiveness and efficiency of our proposed method. We firstly introduce the benchmark datasets and evaluation metrics used in Section \ref{dataset} . Then, we describe the training protocol for our experiments in detail in Section \ref{protocal}. After that, we perform ablation study to evaluate each component of our proposed method in Section \ref{ablation}. Finally, we compare the proposed method with several state-of-the-art semantic segmentation methods on the Cityscapes and CamVid datasets in Section \ref{state} and Section \ref{CamVid}, respectively.

\subsection{Datasets and Evaluation Metrics}
\label{dataset}

In this paper, the Cityscapes and CamVid datasets are used for performance evaluation.

The Cityscapes dataset is one of the most challenging semantic segmentation benchmarks. This dataset is large with fine pixel-wise semantic annotations, and it focuses on the understanding of urban street scenes. It contains $5000$ high-resolution images with the size of $1024\times2048$ across 50 cities, different seasons and varying background from a car perspective, which is a formidable challenge for the real-time semantic segmentation task.

The Cityscapes dataset is splited into three parts: training, validation and test datasets, with 2975, 500 and 1525 images, respectively. According to the content of scenes, the dataset is usually annotated into 30 semantic categories. However, like state-of-the-art methods \cite{zhao2017icnet, romera2018erfnet}, 19 common semantic categories (such as road, car and person) are only used for training and evaluation in our experiments. Besides the 5000 finely annotated images, the Cityscapes dataset also provides additional 20000 images with coarse annotations. These images can be used to pretrain the models.
%In this paper, we only use the fine annotations for our experiments.
For the test images, the Cityscapes dataset does not provide the corresponding ground-truth images to the users, but they are stored in the official server and can be used for evaluation online, for fair comparison.

The CamVid dataset is another popular dataset for urban street scene understanding.  CamVid contains 701 densely annotated images extracted from five video sequences
with the resolution of $720\times960$.
For the fair comparison with state-of-the-art methods, all images are divided into 367 for training, 101 for validation and 233 for testing.
The proposed method is trained based on the training and validation datasets, and evaluates the segmentation results on 11 common semantic categories (such as road, car and building) on the training and testing process.

Following other real-time semantic segmentation methods \cite{zhao2017icnet, yu2018bisenet}, we adopt the mean Intersection over Union (mIoU) and frames per second (fps) for single forward propagation as our evaluation metrics, which are used to evaluate the accuracy and speed of all the competing methods. Moreover, the number of parameters (Params) and float-point operations (FLOPs) are also used to evaluate the memory consumption and computational complexity, respectively.

\subsection{Training Protocol}
\label{protocal}
Image resolution is one of the critical factors that affect the inference speed.
Although the larger resolution of the input images leads to improved accuracy, the computational time of the semantic segmentation methods increase proportionally \cite{zhao2017icnet}. Therefore, limiting the image resolution can reduce the complexity of our network and improve the inference speed. In this paper, we downsample the input images to a moderate size.  We resize the input images from $1024\times2048$ to $512\times1024$ for training and $448\times896$ for testing on the Cityscapes dataset. All the images are kept unchanged on the CamVid dataset.

For all the images, we use the commonly-used data augmentation strategies, including horizontal flipping, random scaling (ranging from 0.5 to 0.8) and random cropping, to enlarge the dataset in our training process. Any other sophisticated data augmentation strategies are not used, since we find that they might not be beneficial for our training. For example, the random rotating significantly decreases the accuracy of our method. The reason may be that the lightweight network does not suffer from serious over-fitting problem. Therefore, it is not necessary to use too aggressive data augmentation strategies.

Instead of training from scratch, we use MobileNetV2 pretrained on the ImageNet dataset to initialize the baseline network, and then fine-tune it. Through our experiments, we use the online bootstrapping strategy to alleviate the issue of class imbalance \cite{wu2016high}. We adopt stochastic gradient descent (SGD) \cite{bottou2010large} with an initial learning rate of 0.006 and a weight decay of 0.0005 for training.
We set the momentum to 0.9. The batch size is set to 16. Similar to other semantic segmentation methods, we adopt the popular 'poly' learning rate strategy \cite{liu2015parsenet} (the initial learning rate is multiplied by ${(1-\frac{iter}{max\_iter})}^{power}$ for each iteration with $power = 0.9$, where $iter$ and $max\_iter$ denote the current number of iterations and the maximum number of iterations, respectively) instead of the 'step' strategy (the initial learning rate is decreased at the fixed steps).

All of our models are trained for 200 epochs. The implementations are built on PyTorch. We train and evaluate these models using only a single NVIDIA TITAN X card under CUDA 9 and CUDNN 6. And note that we use some tips and compression in the experiments.

\subsection{Ablation Study}
\label{ablation}
Since the proposed method is composed of four components (LBN-AA, DASPP, SPN and FFN), we investigate and describe the effectiveness of each component step by step in this subsection.
%To be fair, all the experiments are built on the Cityscapes dataset without using any additional datasets (such as MS COCO \cite{lin2014microsoft}).

\subsubsection{Ablation For LBN-AA}

\begin{figure*}[!t]
\centerline{
\includegraphics[width=6.8in]{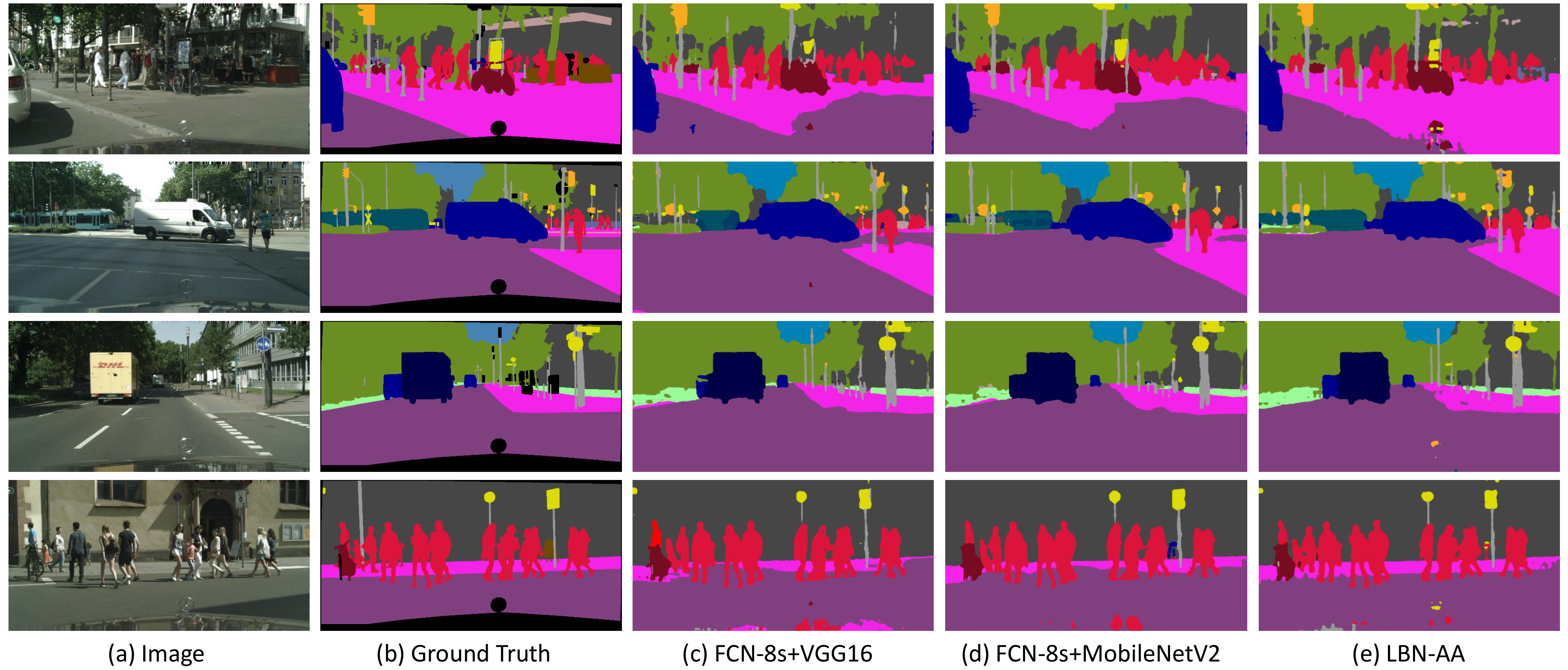}}
\caption{Example results obtained by LBN-AA, VGG16 and MobileNetV2 based on FCN-8s on the Cityscapes validation dataset. Our LBN-AA obtains more detailed information than other baseline networks.}
\label{fig:baseline_compare}
\end{figure*}

\begin{table}[!t]
\renewcommand{\arraystretch}{1.1}
\caption{{The Accuracy and Speed Analysis of Different Baseline Networks: VGG16, MobileNetV2 and LBN-AA on the Cityscapes Validation Dataset.}
% Here, We Use the FCN-8s As the Base Structure.
}
\label{tab:ablation_baseline_network}
\centering
\scalebox{1}{
\begin{tabular}{c|cc} %|c||c|
\Xhline{1.2pt}
{Baseline Network} & {mIoU (\%)} & {Speed (fps)} \\
%\Xhline{1.2pt}
\hline
\hline
FCN-8s+VGG16 & 65.5 & 20.4 \\
FCN-8s+MobileNetV2 & 64.7 & 65.8 \\
LBN-A & 67.5 & 66.8 \\
LBN-ASE & 68.9    & 64.3\\
LBN-AA & \textbf{69.7} & \textbf{64.5} \\
\Xhline{1.2pt}
\end{tabular}
}
\end{table}

We use the modified MobileNetV2 with atrous convolution and attention (LBN-AA) as our feature extraction baseline network. To evaluate the effectiveness and efficiency of LBN-AA, we compare it with the following two different baseline networks, including the original MobileNetV2 and VGG16, where atrous convolution and attention are not used. {We also compare LBN-AA with LBN-A (denotes the modified MobileNetV2 with only atrous convolution) and LBN-ASE (denotes the modified MobileNetV2 with atrous convolution and the SE block). The only difference between LBN-AA and LBN-A is that CAM is adopted or not. LBN-ASE is used to compare the performance difference between the SE block and the proposed CAM.}
All the baseline networks are pretrained on the ImageNet dataset and adopt the structure similar to FCN-8s \cite{long2015fully}. The input images are firstly downsampled to a uniform size ($448\times896$). Then, we feed the images into above three baseline networks and upsample the final output feature maps to the size of original input images. For MobileNetV2 and VGG16, we directly combine the feature maps which are 1/8, 1/16 and 1/32 of the original input image size by skip connections. Here, bilinear interpolation is used to ensure the same size (1/8) of all feature maps.
%The input images are firstly downsampled to  a uniform size ($400\times800$ is adopted) for testing. Then, we feed the images into the above baseline networks and directly upsample the output feature maps to the size of original images.
We evaluate these baseline networks on the Cityscapes validation dataset and report the results in Table \ref{tab:ablation_baseline_network}. Besides, Fig.~\ref{fig:baseline_compare} presents some visual examples obtained by these different baseline networks.

From Table \ref{tab:ablation_baseline_network}, we can see that the accuracy obtained by LBN-AA is improved by about 4\% mIoU, while the speed is about 3 times faster than FCN-8s+VGG16. Compared with FCN-8s+MobileNetV2, LBN-AA also achieves higher accuracy and competitive speed. %, which proves the validity of LBN-AA.
Therefore, the proposed LBN-AA not only greatly improves the performance in accuracy, but also achieves an astonishing speed in the inference time. In contrast to LBN-A, the accuracy of LBN-AA is improved by about 2\% mIoU, which shows the importance of our proposed CAM.
{Compared with LBN-ASE, LBN-AA achieves better segmentation accuracy. The SE block can improve the representational power of the network by modelling channel-wise relationships. However, the ReLU and linear operation used in the SE block might not optimal for our LBN, since the information flow in the lightweight network is restricted. In contrast, the proposed CAM introduces Leaky ReLU and $1 \times 1$ Convolution to respectively extract more informative features and increase the nonlinearity. Such a manner is important for the lightweight network.}
From Fig.~\ref{fig:baseline_compare}, we can see that LBN-AA obtains more detailed information (e.g., some trucks, roads or humans) than other baseline networks. % while *** lose *** and misclassify **.
%This is mainly because that all convolution and pooling layers after the last bottleneck layer of MobileNetV2 are removed in LBN-AA, thus improving the efficiency. Moreover,  atrous convolution and feature concatenation are beneficial to improve the accuracy [please confirm].
This demonstrates the effectiveness of our proposed LBN-AA. In the following experiments, we use LBN-AA as our baseline network.

\subsubsection{Ablation For DASPP}

\begin{table}[!t]
\renewcommand{\arraystretch}{1.1}
\caption{{The Performance Comparison Between ASPP and DASPP on the Cityscapes Validation Dataset.}}
% Here, LBN-AA is Used As the Baseline Network for This Experiment.
\label{tab:ablation_ASPP}
\centering
\scalebox{0.95}{
\begin{tabular}{c|cc} %|c||c|
\Xhline{1.2pt}
 {Method} & {mIoU (\%)} & {Speed (fps)} \\
%\Xhline{1.2pt}
\hline
\hline
LBN-AA+ASPP & 70.8 & 58.3 \\
LBN-AA+DASPP\_Max-pooling & 71.3 & 56.0 \\
LBN-AA+DASPP\_Avg-pooling & \textbf{72.2} & \textbf{56.6} \\
LBN-AA+DASPP (sum) & 71.2 & 56.6 \\
LBN-AA+Vortex Pooling & 71.5 & 51.3 \\
\Xhline{1.2pt}
\end{tabular}
}
\end{table}

\begin{figure*}[!t]
\centerline{
\includegraphics[width=6.8in]{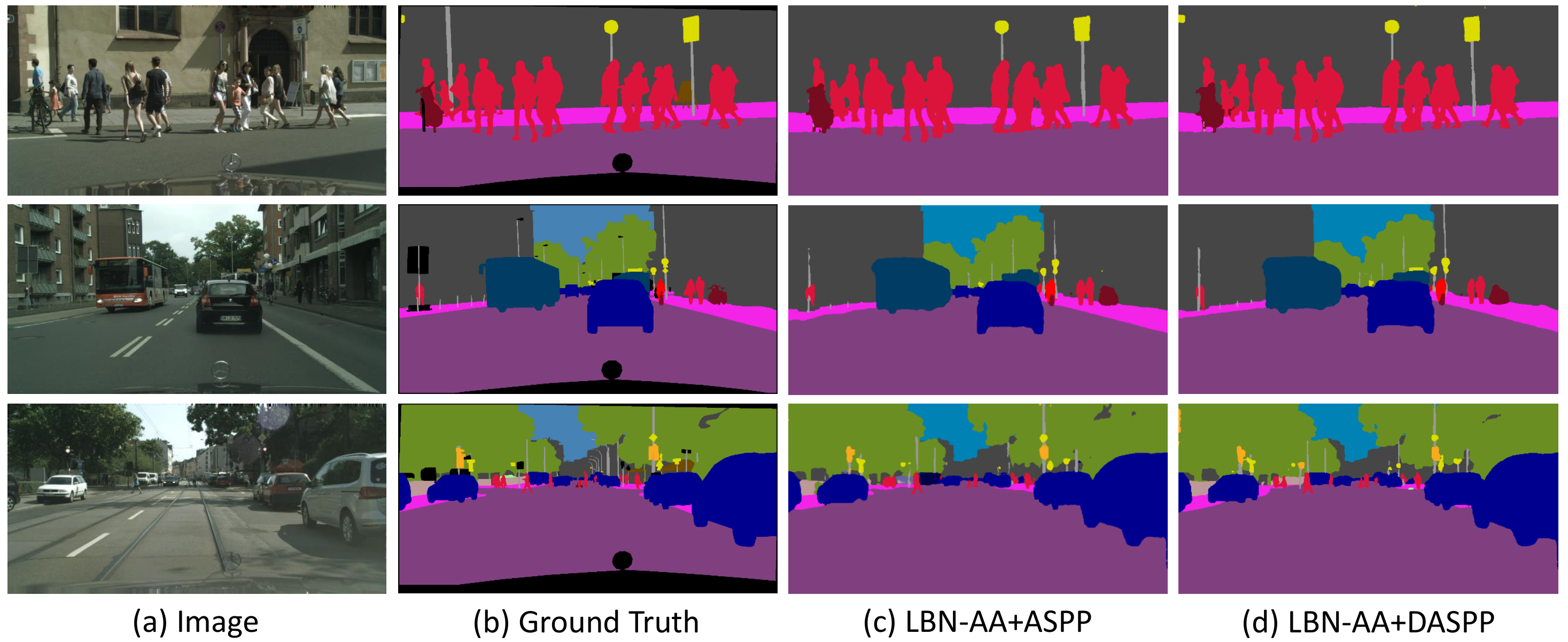}}
\caption{{Qualitative segmentation results obtained by LBN-AA+ASPP and LBN-AA+DASPP (average pooling is used) on the Cityscapes validation dataset.}}
\label{fig:aspp_compare}
\end{figure*}

%In semantic segmentation tasks, ASPP is widely applied to deal with the multi-scale problem. The disadvantages of previous ASPP are that each parallel branch directly convolutes the same feature maps from the former layer, which causes that the result features are less distinctive, and each convolution only uses the information of feature maps at non-zero points of filters. Therefore, we add an average pooling operation with different sizes before each atrous convolution branch and make some other modifications to overcome the shortcomings.

To demonstrate the effectiveness of DASPP, we compare the performance obtained by ASPP \cite{chen2017rethinking} and our proposed DASPP based on the above LBN-AA.
The experimental comparison is shown in Table \ref{tab:ablation_ASPP}. LBN-AA+ASPP denotes that we combine LBN-AA and ASPP in cascade, while LBN-AA+DASPP denotes that we combine LBN-AA and DASPP. Here, two different types of pooling operations, including the max pooling (LBN-AA+DASPP\_Max-pooling) and the average pooling (LBN-AA+DASPP\_Avg-pooling) used in the pooling layers, are respectively evaluated for comparison. We also evaluate the performance of DASPP with the connection mode that directly performs the element-wise addition of the shortcut connection and the outputs from  the four convolution branches (denoted as LBN-AA+DASPP (sum)). In addition,  we evaluate the performance of LBN-AA+Vortex Pooling, which denotes that we combine LBN-AA and the vortex pooling  \cite{xie2018vortex}.
Fig.~\ref{fig:aspp_compare} presents some qualitative segmentation examples obtained by these different models on the Cityscapes validation dataset.

%we also compare the performance of DASPP with different connection modes.

From Table \ref{tab:ablation_ASPP}, we can observe that DASPP boosts the mIoU from 70.8\% (obtained by ASPP) to 71.3\% (max pooling) and 72.2\% (average pooling).
The proposed DASPP brings an improvement in accuracy, while slightly decreasing the computational speed compared with ASPP. The disadvantages of ASPP are that each parallel branch directly adopts the same feature maps from the former layer, which leads to the problem that the obtained features are less distinctive, and each atrous convolution only exploits the information of feature maps at non-zero positions of filters. Compared with LBN-AA+DASPP\_Max-pooling, LBN-AA+DASPP\_Avg-pooling achieves higher mIoU. This is mainly because the average pooling makes use of the information from all the pixels to obtain the feature maps for each atrous convolution branch. In contrast, the max pooling only considers one pixel, which might not be beneficial for semantic segmentation (e.g., the pixels from the boundary may disturb the classification).
In the following, we use the average pooling in DASPP.

LBN-AA+DASPP\_Avg-pooling obtains about 1\% mIoU higher than LBN-AA+DASPP (sum) while achieving the same inference speed.
This validates the effectiveness of the connection mode used in DASPP, where the outputs from the four atrous convolution branches are concatenated, and then the element-wise addition of the shortcut connection and the concatenated outputs is calculated. Furthermore, compared with LBN-AA+Vortex Pooling,  LBN-AA+DASPP achieves better accuracy and faster speed, which demonstrates the superiority of the network architecture of DASPP. {The vortex pooling performs the element-wise addition of the outputs from different atrous convolution branches. However, the simple addition operation may lose some critical information in some branches. In contrast, DASPP concatenates the outputs from different branches, and then performs the element-wise addition of the shortcut connection and the concatenated outputs. In this way, DASPP is able to more effectively exploit the discriminative information from different branches in comparison with
the vortex pooling.}

%From Fig.~\ref{fig:aspp_compare}, the different sizes of objects can be more effectively distinguished by DASPP in a complex urban street scene compared with other methods. This is due to the fact that DASPP increases the diversity of input features for the atrous convolution operations of all the branches, and takes advantage of more contextual information using the different sizes of pooling operations.

\subsubsection{Ablation For SPN}

\begin{table}[!t]
\renewcommand{\arraystretch}{1.1}
\caption{The Accuracy and Speed Comparison Between LBN-AA+DASPP and LBN-AA+DASPP+SPN on the Cityscapes Validation Dataset.}
\label{tab:ablation_branch}
\centering
\scalebox{1}{
\begin{tabular}{c|cc} %|c||c|
\Xhline{1.2pt}
{Method} & {mIoU (\%)} & {Speed (fps)} \\
%\Xhline{1.2pt}
\hline
\hline
LBN-AA+DASPP & 72.2 & 56.6 \\
LBN-AA+DASPP+SPN & \textbf{73.6} & \textbf{53.5} \\
\Xhline{1.2pt}
\end{tabular}
}
\end{table}

In this subsection, we evaluate the importance of our SPN for semantic segmentation.
The results are listed in Table \ref{tab:ablation_branch}.
LBN-AA+DASPP+SPN denotes that we combine LBN-AA+DASPP and SPN in parallel, which effectively encodes the semantic and spatial information, respectively. Note that the commonly-used element-wise addition strategy is used to fuse the feature maps from LBN-AA+DASPP and SPN. %Figure [**************] presents some visual examples obtained by these different models (including LBN-AA+DASPP and LBN-AA+DASPP+SPN) on the Cityscapes validation dataset.

From Table \ref{tab:ablation_branch}, we can see that the accuracy obtained by LBN-AA+DASPP+SPN is improved by 1.4\% mIoU while only slightly decreasing the speed. LBN-AA+DASPP employs the deep network to improve the capacity of feature extraction and provide large receptive fields. However, it inevitably suffers from the serious challenge of losing much spatial information.
In contrast, SPN effectively preserves the rich spatial detail information, such as small traffic signs or objects in the distance. In other words, it provides the spatial information complementary to the semantic information given by LBN-AA+DASPP. This demonstrates the importance of SPN.

%With the development of deep learning, many state-of-the-art methods

%With the development of deep learning, many modern approaches utilize the special deep network frameworks to improve the capacity of feature extraction and provide larger receptive fields. However, they also inevitably suffer from the serious challenge of lost of spatial information. Therefore, we propose an auxiliary refining branch to preserve the rich spatial detail information and capture the large size of feature maps. Based on the ablation studies above, we add this branch to our best model to conduct experiments. Note that for the network with our auxiliary refining branch . We evaluate the performance of above best model with the auxiliary refining branch and without the branch respectively.

\subsubsection{Ablation For FFN}

%To further improve the performance of our method, we specifically design a feature fusion network (FFN) to effectively combine the features from the LBN-AA+DASPP and SPN.
In this subsection, we evaluate the effectiveness of our proposed FFN for feature fusion.
As we mention above, LBN-AA+DASPP and SPN respectively encode the semantic and spatial information. Therefore, we compare different fusion strategies, including the commonly-used element-wise addition (abbreviated as Ele-wise Add) and the proposed FFN. The results are reported in Table \ref{tab:ablation_module}. %Figure [**************] presents some visual examples obtained by these different methods (including LBN-AA+DASPP+SPN and the proposed method) on the Cityscapes validation dataset.

Although the element-wise addition of the features is more efficient than our proposed method (using FFN), the performance is lower than ours (the accuracy of the proposed method is improved by 0.8\% mIoU). This is mainly because that FFN can effectively fuse the deep (semantic) and shallow (spatial) features from the different levels.
Note that our proposed method can still achieve the real-time inference speed using FFN.
\begin{table}[!t]
\renewcommand{\arraystretch}{1.2}
\caption{{The Accuracy and Speed Comparison of Different Fusion Strategies on the Cityscapes Validation Dataset.}}
\label{tab:ablation_module}
\centering
\scalebox{0.82}{
\begin{tabular}{cc|cc} %|c||c|
\Xhline{1.2pt}
{Method} & {Fusion Strategy} & {mIoU (\%)} & {Speed (fps)} \\
%\Xhline{1.2pt}
\hline
\hline
LBN-AA+DASPP+SPN & Ele-wise Add & 73.6 & 53.5 \\
Ours & FFN & \textbf{74.4} & \textbf{51.0} \\
\Xhline{1.2pt}
\end{tabular}
}
\end{table}
%Therefore, the above ablation study shows that each component of the proposed method is effective and efficient.
\subsection{Comparison with State-of-the-art Methods}

%Here we firstly combine the above four components into a complete network and make a final fine-tuning to obtain our best network framework. Then we evaluate this best model on the validation set of Cityscapes and list the results in Table \ref{tab:validation_comparison}.

%Here we firstly combine the above four components into a complete network and make a fine-tuning. To further improve feature representation, we also add the latest dual attention module \cite{fu2018dual} behind the DASPP at the expense of some time, which models the semantic interdependencies in spatial and channel dimensions, to obtain our best network framework.

We firstly evaluate several representative real-time semantic segmentation methods, including the simplified PSPNet \cite{zhao2017pyramid}, ICNet \cite{zhao2017icnet} and our proposed method on the Cityscapes validation dataset. The simplified PSPNet is compressed with the kernel keeping rate of 0.5. ICNet is the recent promosing real-time semantic segmentation method. The comparison results are reported in Table \ref{tab:validation_comparison}.

We can see that the proposed method obtains much better performance than ICNet and the simplified PSPNet in both accuracy and speed. Specifically, the mIoU obtained by the proposed method is about 6.5\% higher and the speed is faster than ICNet. The proposed method achieves an impressive result, which demonstrates the effectiveness and efficiency of our real-time high-performance semantic segmentation method.

Finally, we compare the proposed method with several state-of-the-art methods.
%on the Cityscapes test dataset.
%Note that different from ICNet, which is trained both on the training and validation datasets, our method only uses the training dataset for training.
In Table \ref{tab:testing_comparison}, we report
the accuracy results (mIoU), the corresponding inference speed, running time as well as the status of Params and FLOPs obtained by all the competing methods on the Cityscapes test dataset.
%we report the FLOPs, the number of parameters (\#Params), running time, the inference speed (fps) as well as the corresponding accuracy results (mIoU)
%The status of Params and are also listed for comparison in the table, which indicates the number of operations to process images.
The performance-oriented PSPNet and DeepLabv2 are also used for comparison. Note that all the competing methods are evaluated by using a single NVIDIA TITAN X card.

\begin{table}[!t]
\renewcommand{\arraystretch}{1}
\caption{{The Accuracy and Speed Comparison Between the Proposed Method and PSPNet, Fast ICNet on the Cityscapes Validation Dataset.}}
%Note That the PSPNet is a Simplified Version, Which is Compressed with the Kernel Keeping Rate of 0.5.
\label{tab:validation_comparison}
\centering
\scalebox{1}{
\begin{tabular}{l|cccc} %|c||c|
\Xhline{1.2pt}
{Item} & {PSPNet} & {ICNet} & {Ours} \\
%\Xhline{1.2pt}
\hline
\hline
mIoU (\%) & 67.9 & 67.7 & \textbf{74.4} \\
Time (ms) & 170 & 33 & \textbf{20} \\
Speed (fps) & 5.9 & 30.3 & \textbf{51.0} \\
Image size & $713\times713$   & $1024\times2048$  & $448\times896$\\
\Xhline{1.2pt}
\end{tabular}
}
\end{table}

\begin{table*}[!t]
\renewcommand{\arraystretch}{1}
\centering
\caption{{Comparisons Between the Proposed Method and Other State-of-the-Art Methods on the Cityscapes Test Dataset.
%Sorted by Increasing Mean IoU.
%These Reported Mean IoU and Inference Time are from the Official Leadboard.
Our Final Results are in Boldface. ``-" Indicates that the Corresponding Result is not Provided by the Methods.}}
\label{tab:testing_comparison}
\scalebox{0.95}{
\begin{tabular}{l|c|c|c|c|c|c} %|c||c|
\Xhline{1.2pt}
{Method} & {Input Size} & {FLOPs (G)} & {Params (M)} & {Time (ms)} & {Speed (fps)} & {{mIoU (\%)}} \\
%\Xhline{1.2pt}
\hline
\hline
DeepLabv2 \cite{chen2018deeplab} & $512\times1024$ & 457.8 & 262.1 & 4000 & 0.25 & 63.1 \\
PSPNet \cite{zhao2017pyramid} & $713\times713$ & 412.2 & 250.8 & 1288 & 0.78 & 81.2 \\
\hline
SegNet \cite{badrinarayanan2015segnet} & $640\times360$ & 286 & 29.5 & 60 & 16.7 & 57 \\
ENet \cite{paszke2016enet} & $640\times360$ & 3.8 & 0.4 & 7 & 135.4 & 57 \\
SQNet \cite{treml2016speeding} & $1024\times2048$ & 270 & - & 60 & 16.7 & 59.8 \\
CRF-RNN \cite{zheng2015conditional} & $512\times1024$ & - & - & 700 & 1.4 & 62.5 \\
%DeepLabv1 \cite{chen2014semantic} & $\times$ &  &  &  &  &  \\
FCN-8S \cite{long2015fully} & $512\times1024$ & 136.2 & - & 500 & 2.0 & 63.1 \\
FRRN \cite{pohlen2017fullresolution} & $512\times1024$ & 235 & - & 469 & 2.1 & 71.8 \\
ERFNet \cite{romera2018erfnet} & $512\times1024$ & 27.7 & 2.1 & 24 & 41.7 & 69.7 \\
%Adelaide \cite{lin2016efficient} & $\times$ &  &  &  &  &  \\
%Dilation10 \cite{yu2015multi} & $\times$ &  &  &  &  &  \\
ICNet \cite{zhao2017icnet} & $1024\times2048$ & 28.3 & 26.5 & 33 & 30.3 & 70.6 \\
TwoColumn \cite{wu2017real} & $512\times1024$ & 57.2 & - & 68 & 14.7 & 72.9 \\
SwiftNetRN \cite{orsic2019defense} & $1024\times2048$ & 114.0 & 12.9 & 56 & 18.0 & 75.1 \\
%ERFNet \cite{romera2018erfnet} & 69.7 & 24 & 41.6 \\
%\textbf{Our} & \textbf{69.7} & \textbf{26.3} & \textbf{38.0} \\
%GUNet \cite{mazzini2018guided} & 70.4 & 30 & 33.3 \\
%DeepLabv2-CRF \cite{chen2018deeplab} & 70.4 & n/a & n/a \\
LEDNet \cite{wang2019lednet} & $512\times1024$ & - & 0.94 & 25 & 40 & 70.6 \\
BiSeNet1 \cite{yu2018bisenet} & $768\times1536$ & 14.8 & 5.8 & 14 & 72.3 & 68.4 \\
BiSeNet2 \cite{yu2018bisenet} & $768\times1536$ & 55.3 & 49 & 22 & 45.7 & 74.7 \\
DFANet \cite{li2019dfanet} & $1024\times1024$ & 3.4 & 7.8 & 10 & 100.0 & 71.3 \\
%DFANet \cite{li2019dfanet} & $512\times1024$ & 1.7 & 7.8 & 6 & 160 & 70.3 \\
\hline
\hline
\textbf{Ours} & \textbf{$448\times896$} & \textbf{49.5} & \textbf{6.2} & \textbf{20} & \textbf{51.0} & \textbf{73.6} \\
\Xhline{1.2pt}
\end{tabular}
}
\end{table*}

\begin{table*}[!t]
\renewcommand{\arraystretch}{1}
\centering
\caption{{The Per-class, Class and Category IoU Evaluation Results on the Cityscapes Test Dataset.
List of Classes (From Left to Right): Road, Side-Walk, Building, Wall, Fence, Pole, Traffic Light, Traffic Sign, Vegetation, Terrain, Sky, Pedestrian, Rider, Car, Truck, Bus, Train, Motorbike and Bicycle. ``Cla" Denotes mIOU (19 Classes), ``Cat" Denotes mIOU (7 Categories).
Our Final Results are in Boldface.}}
\label{tab:testing_classwise}
\scalebox{0.74}{
\begin{tabular}{l||ccccccccccccccccccc||cc} %|c||c|
\Xhline{1.2pt}
{Method} & {Roa} & {Sid} & {Bui} & {Wal} & {Fen} & {Pol} & {TLi} & {TSi} & {Veg} & {Ter} & {Sky} & {Per} & {Rid} & {Car} & {Tru} & {Bus} & {Tra} & {Mot} & {Bic} & {Cla} & {Cat} \\
\hline
\hline
ENet \cite{paszke2016enet} &96.3&74.2&85.0&32.2&33.2&43.5&34.1&44.0&88.6&61.4&90.6&65.5&38.4&90.6&36.9&50.5&48.1&38.8&55.4&58.3&80.4 \\
FCN-8S \cite{long2015fully}  &97.4&78.4&89.2&34.9&44.2&47.4&60.1&65.0&91.4&69.3&93.9&77.1&51.4&92.6&35.3&48.6&46.5&51.6&66.8&65.3&85.7 \\
ERFNet \cite{romera2018erfnet} &97.9&82.1&90.7&45.2&50.4&59.0&62.6&68.4&91.9&69.4&94.2&78.5&59.8&93.4&52.3
&60.8&53.7&49.9&64.2&69.7&87.3 \\
LEDNet \cite{wang2019lednet} &98.1&79.5&91.6&47.7&49.9&62.8&61.3&72.8&92.6&61.2&94.9&76.2&53.7&90.9&64.4&64.0&52.7&44.4&71.6&70.6&87.1 \\
\hline
\hline
\textbf{Ours}&\textbf{98.2}&\textbf{84.0}&\textbf{91.6}&\textbf{50.7}&\textbf{49.5}&\textbf{60.9}&\textbf{69.0}&\textbf{73.6}&\textbf{92.6}&\textbf{70.3}&\textbf{94.4}&\textbf{83.0}&\textbf{65.7}&\textbf{94.9}&\textbf{62.0}&\textbf{70.9}&\textbf{53.3}&\textbf{62.5}&\textbf{71.8}&\textbf{73.6}&\textbf{88.8} \\
\Xhline{1.2pt}
\end{tabular}
}
\end{table*}

From Table \ref{tab:testing_comparison}, we can observe that our proposed method achieves significant progress on both
 speed and accuracy. Specifically, the proposed method obtains the performance of 73.6\% mIoU at the speed of 51.0 fps, and requires only 49.5G FLOPs, 6.2M Params.

The proposed method is faster and more accurate than most of the state-of-the-art methods, such as ERFNet, ICNet, BiSeNet1. The proposed method is about 16\% mIoU higher than the fast-speed ENet. Although ENet, a representative real-time segmentation method, is more efficient (in terms of FLOPs, Params and speed) than the proposed method, it obtains much worse segmentation accuracy. The proposed method is even better than some accuracy-oriented methods, such as DeepLabv2 (i.e., it is about 200 times faster and 10\% mIoU higher compared with the DeepLabv2). Although the proposed method is slightly slower than LEDNet and DFANet, it is also real-time and is much more accurate. LEDNet and DFANet show poor segmentation accuracy (they obtain more than 2\% mIoU drop than the proposed method). Compared with BiSeNet2 and SwiftNetRN, the proposed method is faster and provides a competitive accuracy, while it requires less computational complexity and memory consumption.  {BiSeNet2 adopts the two-path framework, where the memory consumption is relatively high
(e.g., about 49M Params). In this paper, we also use the two-path framework,
where two simple modules (i.e., SPN and FFN) are developed. The components used in our
method are totally different from the modules used in BiSeNet2. As a result, the proposed method requires only 6.2M Params (about 8 times smaller than BiSeNet2).}
Therefore, the proposed method achieves an excellent tradeoff between the accuracy and speed among all the competing methods.
%great progress against the other methods both in accuracy and speed. As can be observed, our method makes a best tradeoff between accuracy and speed among all the methods.

In Table \ref{tab:testing_classwise}, we present the detailed per-class, class and category IoU results on the Cityscapes test dataset. The proposed method  achieves the best accuracy by significant margins on most of the classes, while keeping a faster
speed than the other competing methods. This demonstrates the effectiveness and efficiency of the proposed method.

\begin{figure*}[!t]
\centerline{
\includegraphics[width=6in]{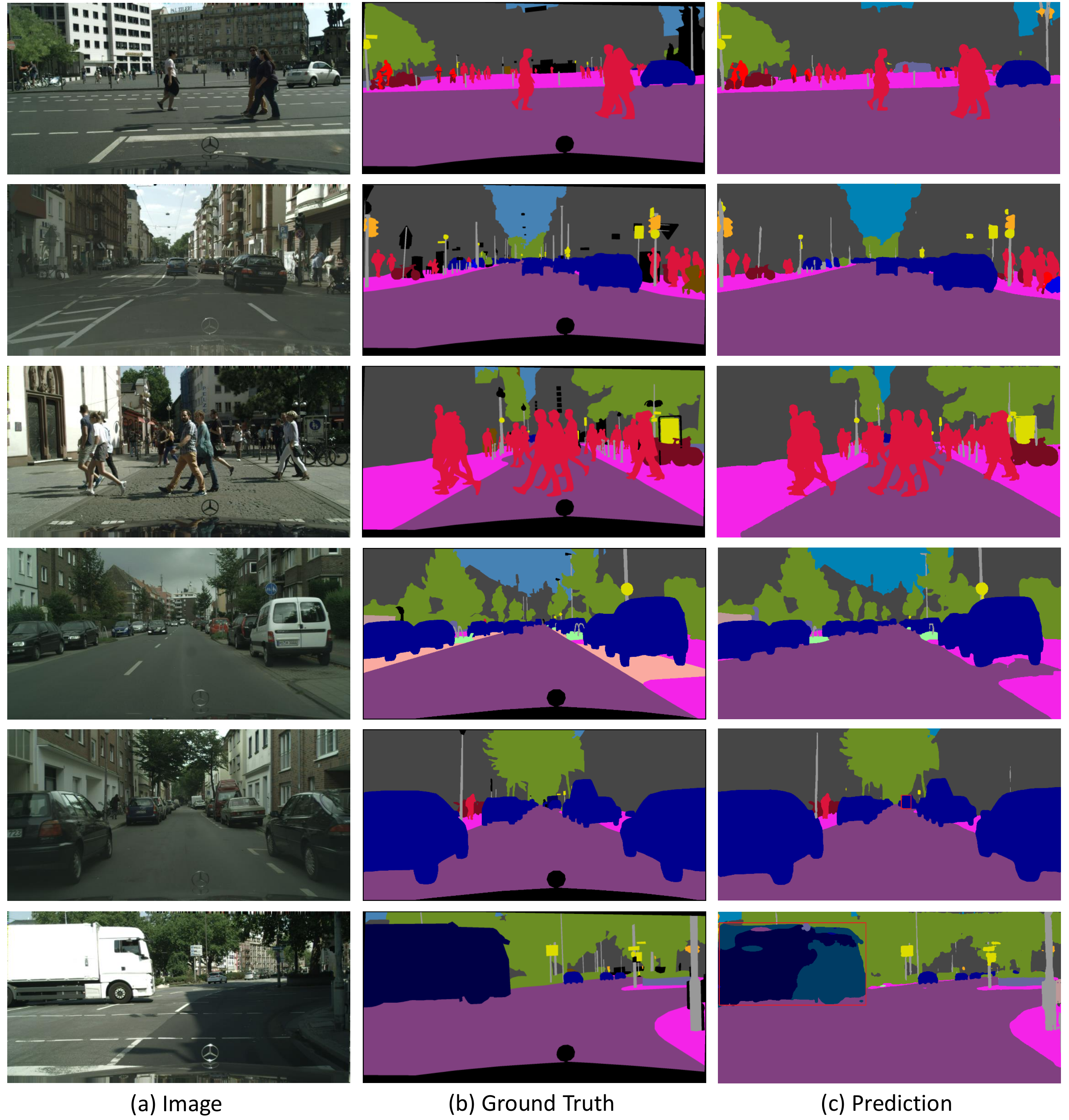}}
\caption{{Visualization results on the Cityscapes validation dataset. From left to right respectively input images, ground-truth images and our predicted results. The last two rows show some failure cases.}}
\label{fig:segmentation_results}
\end{figure*}

We also provide the qualitative visual results including the successful and failing predictions, as shown in Fig.~\ref{fig:segmentation_results}. Generally speaking, our proposed method can correctly assign the labels to different objects (see the first four rows in Fig.~\ref{fig:segmentation_results}) in the complex urban street scenes (suffering from multi-scale variations, illumination changes, occlusions, etc.). However, it might sometimes over-segment the area inside the objects or under-segment between the objects (e.g., the truck in the last row of Fig.~\ref{fig:segmentation_results}). The main reason might be that the appearance inside the objects is significantly different, thus leading to the over-segmenting problem. Under-segmenting is the opposite.
Note that this problem also exists for other semantic segmentation methods, such as ICNet.

\label{state}

\subsection{Comparison on the CamVid Dataset}
\label{CamVid}
To illustrate the generality and effectiveness of our proposed method, we also evaluate our method on the CamVid dataset, which contains images extracted from the video sequences with resolution up to $720\times960$. The evaluation results are reported in Table \ref{tab:testing_camvid}. It can be seen that our proposed method is comparable to the state-of-the-art BiSeNet in accuracy. Although the speed of our proposed method is slightly slower than DFANet, our method achieves much better segmentation accuracy.  {In addition, to further show the strength and limitations of our proposed method, we give the detailed statistic accuracy results in Table \ref{tab:testing_camvid_classwise}. We can see that the proposed method is able to more accurately segment small objects (such as sign, pole) compared with other methods. This is mainly because the effectiveness of the proposed semantic branch (LBN+AA and DASPP) and the spatial branch (SPN), which respectively encode distinctive semantic information and detailed spatial information.
However, the proposed method achieves worse segmentation accuracy than other competing methods for the class of tree. The reason is that the appearance inside the tree can be significantly different and is very similar to the background. As a result, the proposed method over-segments the area inside the tree.
In a word, the proposed method again achieves the outstanding performance in both accuracy and speed.}
%Therefore, our method still achieves a good balance between accuracy and speed on the CamVid dataset.}

\begin{table}[!t]
\renewcommand{\arraystretch}{1}
\caption{{The Accuracy and Speed Comparisons of the Proposed Method against Other Methods on the CamVid Test Dataset.
%Sorted by Increasing Mean IoU.
Our Final Results are in Boldface.}}
\label{tab:testing_camvid}
\centering
\scalebox{1}{
\begin{tabular}{l|c|c|c} %|c||c|
\Xhline{1.2pt}
{Method} & {Time (ms)} & {Speed (fps)} & {mIoU (\%)} \\
%\Xhline{1.2pt}
\hline
\hline
SegNet \cite{badrinarayanan2015segnet} & 217 & 46 & 46.4 \\
DPN \cite{liu2015semantic} & 830 & 1.2 & 60.1 \\
DeepLabv2 \cite{chen2018deeplab} & 203 & 4.9 & 61.6 \\
ENet \cite{paszke2016enet} & - & - & 51.3 \\
ICNet \cite{zhao2017icnet} & 36 & 27.8 & 67.1 \\
BiSeNet1 \cite{yu2018bisenet} & - & - & 65.6 \\
BiSeNet2 \cite{yu2018bisenet} & - & - & 68.7 \\
DFANet \cite{li2019dfanet} & 8 & 120 & 64.7 \\
\hline
\hline
\textbf{Ours} & \textbf{25} & \textbf{39.3} & \textbf{68.0} \\
\Xhline{1.2pt}
\end{tabular}
}
\end{table}

\begin{table*}[!t]
\renewcommand{\arraystretch}{1}
\centering
\caption{{The Detailed Statistic Accuracy Results on the CamVid Test Dataset. List of Classes (From Left to Right): Building, Tree, Sky, Car, Sign, Road, Pedestrain, Fence, Pole, Sidewalk and Bicyclist. ``Cla" Denotes mIOU (11 Classes). Our Final Results are in Boldface.}}
\label{tab:testing_camvid_classwise}
\scalebox{0.92}{
\begin{tabular}{l||ccccccccccc||cc} %|c||c|
\Xhline{1.2pt}
{Method} & {Bui} & {Tre} & {Sky} & {Car} & {Sig} & {Roa} & {Ped} & {Fen} & {Pol} & {Sid} & {Bic} & {Cla} \\
\hline
\hline
SegNet \cite{badrinarayanan2015segnet} &88.8&87.3&92.4&82.1&20.5&97.2&57.1&49.3&27.5&84.4&30.7&55.6 \\
ENet \cite{paszke2016enet} &74.7&77.8&95.1&82.4&51.0&95.1&67.2&51.7&35.4&86.7&34.1&51.3 \\
BiSeNet1 \cite{yu2018bisenet} &82.2&74.4&91.9&80.8&42.8&93.3&53.8&49.7&25.4&77.3&50.0&65.6 \\
BiSeNet2 \cite{yu2018bisenet} &83.0&75.8&92.0&83.7&46.5&94.6&58.8&53.6&31.9&81.4&54.0&68.7 \\
\hline
\hline
\textbf{Ours}&\textbf{83.2}&\textbf{70.5}&\textbf{92.5}&\textbf{81.7}&\textbf{51.6}&\textbf{93.0}&\textbf{55.6}&\textbf{53.2}&\textbf{36.3}&\textbf{82.1}&\textbf{47.9}&\textbf{68.0} \\
\Xhline{1.2pt}
\end{tabular}
}
\end{table*}

\section{Conclusion}
\label{conclusion}
In this paper, we propose a novel real-time high-performance semantic segmentation method to achieve a great tradeoff between accuracy and speed. The proposed method consists of four main components: LBN-AA, DASPP, SPN and FFN. LBN-AA utilizes the lightweight network, atrous convolution, convolutional attention module to efficiently extract features and obtain dense feature maps. DASPP increases the diversity of the input features and takes advantage of rich contextual information to effectively deal with the multi-scale problem of semantic segmentation. SPN is designed to preserve the abundant spatial information and remedy the loss of details. FFN is responsible for fusing the high-level and low-level features.
These components are tightly coupled and jointly optimized to ensure the performance of semantic segmentation.
Both qualitative and quantitative results on the Cityscapes and CamVid datasets demonstrate the effectiveness and efficiency of our proposed method. Besides, we believe that some modules in the proposed method can not only be used for real-time semantic segmentation, but also be used for accuracy-oriented semantic segmentation.

\ifCLASSOPTIONcaptionsoff
  \newpage
\fi

\end{document}